\documentclass[sigconf]{acmart}

\usepackage{graphicx}
\usepackage{amsmath}
\usepackage{subcaption}
\usepackage{epstopdf}
\usepackage{amsthm}
\usepackage{booktabs}
\usepackage{hyperref}

\usepackage{algorithm}
\usepackage{algorithmic}

\usepackage{multirow}

\setcopyright{none}

\acmDOI{00.000/000_0}

\acmISBN{000-0000-00-000/00/00}

\acmConference[0000]{000}{0000}{Beijing, China} 
\acmYear{0000}
\copyrightyear{0000}

\acmPrice{00.00}

\begin{document}
	\title{Topic Model Supervised by Understanding Map}
	
	\author{Gangli Liu}
	\affiliation{%
		\institution{Tsinghua University}
		\city{Beijing} 
		\state{China} 
		\postcode{100084}
	}
	\email{gl-liu13@mails.tsinghua.edu.cn}

\begin{abstract}
	Inspired by the notion of Center of Mass in physics, an extension called Semantic Center of Mass (SCOM) is proposed, and used to discover the abstract ``topic'' of a document. The notion is under a framework model called Understanding Map Supervised Topic Model (UM-S-TM). The devising aim of UM-S-TM is to let both the document content and a semantic network - specifically, Understanding Map - play a role, in interpreting the meaning of a document. Based on different justifications, three possible methods are devised to discover the SCOM of a document. Some experiments on artificial documents and Understanding Maps are conducted to test their outcomes. In addition, its ability of vectorization of documents and  capturing sequential information are tested. We also compared UM-S-TM with probabilistic topic models like Latent Dirichlet Allocation (LDA) and probabilistic Latent Semantic Analysis (pLSA).
\end{abstract}

%
%



\maketitle

\section{Introduction}

People are always interested in summarizing the ``topic'' of a document or a collection of documents. Because the world is exploding with information. Lots of times we need to summarize it to make it informative and simple. One interesting question is: if we input Leo Tolstoy's masterpiece \emph{War and Peace}  into a machine without telling it the title of the book, and ask it to summarize the ``topic''  of the book with two concepts, what would it be?  

To answer this question, we propose a model called Understanding Map Supervised Topic Model (UM-S-TM).

In \cite{Liu2017}, Understanding Map (U-map) is introduced to help people understand complicated concepts in a domain, in a step by step manner. Understanding Map is a type of Semantic Network \cite{woods1975s} that represents semantic relations between concepts in a network.  Unlike a typical semantic network, there are no semantic triples in U-map. Two concepts are connected only if one concept appears in the other's definition. We argue that if a concept appears in another one's definition, there must be some tight semantic bond between the two.  Understanding Map is used to capture these relations, it is simple and easy to construct. 

To construct an Understanding Map, we first set a domain, e.g., Machine Learning, Analytic Geometry, Graph Theory etc. Then compile a glossary of the domain, with definitions of concepts in the glossary. Then construct Understanding Map based on the definitions.

\begin{figure}
	\centering
	\includegraphics[width=0.7\columnwidth]{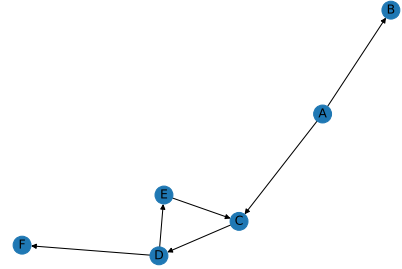}
	\caption{U-map One : a simple artificial Understanding Map which consists of 6 concepts}
	\label{fig:simple-u-map}
\end{figure}

Figure ~\ref{fig:simple-u-map}  is a simple artificial Understanding Map which consists of 6 concepts, with each concept represented by a letter. An edge from Concept A to Concept C means C appears in A's definition. This simple artificial U-map is  used for illustrating logic of UM-S-TM. Real world Understanding Maps are much more complex, containing thousands of or even millions of concepts. In UM-S-TM the directions of edges are not necessary, so they are ignored in following demonstration.

In section 3.2.3 of  \cite{Liu2017}, we mention Understanding Map can be used to illustrate the knowledge characteristics of a person, a corpus, or a document. We re-post Figure 10 of  \cite{Liu2017} - \emph{ A part of a night light map}. Looking at the night light map, imagine it is a document represented by an Understanding Map, with each node's brightness represented by a concept's term frequency in the document. Further, imagine it is a bunch of particles, with each concept being a particle; the mass of the particle is the term frequency of the corresponding concept. The distance between two particles is defined by the distance between two concepts in Understanding Map.

\begin{figure}[h]
	\centering
	\includegraphics[width=0.7\columnwidth]{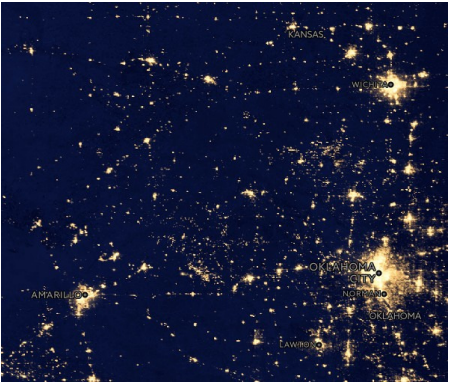}
	\caption{A part of a night light map}
\end{figure}

With this projection - particle to concept, mass to term frequency, physical location to location in Understanding Map, and system of particles to document  - an interesting question is: If we know how to find the Center of Mass of a system of particles, does it exist a counterpart of ``semantic''   Center of Mass to a document? If the answer is \emph{yes}, a natural second question is: What is the ``semantic''   Center of Mass to a document?

\section{Methods of UM-S-TM}  \label{sec-two}
In this section we first formalize some definitions, then propose three possible methods for finding out  the ``semantic''   Center of Mass.
\subsection{Domain, topic, document, and distance}
To find out the Semantic Center of Mass of a document, we formalize the following definitions:

\begin{definition}
Domain

In UM-S-TM, a domain is equivalent to an Understanding Map. It consists of a set of concepts and their relations determined by their definitions.
\end{definition}

 E.g., Figure ~\ref{fig:simple-u-map} is a simple domain that contains six concepts. With techniques of graph partition \cite{Bourse2014, Wang2014}, a domain can be segmented into clusters of subdomains.

\begin{definition}
Topic

In UM-S-TM, a topic is defined as a non-empty set of concepts that is a subset of a domain. 
\end{definition}

Therefore, a domain contains at most $2^n - 1$ topics, where $n$ is the cardinality of the domain. That is the power set of Domain $\Psi$ except the empty set. E.g., Domain of  U-map One  contains 6 concepts, so it contains $2^6 - 1 = 127$ topics.   Following are some example topics in real world:	
\begin{align*} 
	T_1   &= \{war, peace\}    \\
	T_2   &=  \{ guns, roses\}    \\ 
	T_3   &=  \{Latent \  Dirichlet \ Allocation\}    \\
	T_4   &= 	  \{gravitational \ wave, Nelson \  Mandela, Maxim \  gun\}     
\end{align*}
A topic's length is the number of concepts it contains. A concept that belongs to a topic is called a sub-topic of the topic.

\begin{definition}
Document

In UM-S-TM, a document is not presented as a bag-of-words, but a bag-of-concepts. 
\end{definition}
Since we are equipped with Understanding Map, we have a database of concepts in a domain. Using the database to pre-process a document, merging concepts consisting of multi-words, such as  \emph{Latent   Dirichlet Allocation, Gravitational Wave} etc. After the merging, a document is converted to a bag-of-concepts. Two parameters are associated with a document. A document's concept-length calculates how many concept components it contains; a document's length calculates how many tokens it contains.

An n-concept-document means it contains $n$ concept ingredients; it's concept-length is  $n$. Two documents may contain the same set of concepts, but their length may be different due to different term frequencies. E.g., 
\begin{align*} 
	Doc_1   &= \{            A:10, \quad B:0,\quad C:1                         \}    \\
	Doc_2   &=  \{           A:10, \quad B:0, \quad C:99                            \}     
\end{align*}
Document 1 and 2 both are composed of Concept A and C, so their concept-length are two. They have different length. Doc1's length is 11; Doc2's length is 109. A document's Ingredient Topic is defined to be the topic that contains all the concepts the document contains. Therefore, both $Doc_1$ and $Doc_2$'s Ingredient Topic are $\{A,C\}$.

\begin{definition}
Concept-to-concept distance

A concept's distance to another concept is the length of the shortest path between the two, in an Understanding Map. 
\end{definition}
E.g., in U-map One, Concept B's distance to D is 3, note the directions are ignored. Since an U-map is a connected graph, the distance between any two concepts is finite. 

\begin{definition}
Concept-to-topic distance
	
A concept's distance to a topic is defined as the shortest distance from the concept to any concepts in the topic.

\begin{equation}
d(c, \Phi) = \min_{ c_i \in \Phi} d(c, c_i)   
\label{eq:concept-to-topic}
\end{equation}

where $c$ is a concept, $\Phi$ is a topic, $c_i $ is a concept belongs to $\Phi$.
\end{definition}

For example, in U-map One, Concept B's distance to Topic $\{A, E, F\}$ is one, since the shortest distance is from B to A, that is one. 

The topic that contains all concepts in a domain is an uninformative topic, because all concepts' distance to this topic is zero. So it is excluded when considering all the topics in a domain, as shown in Table ~\ref{tab:appA1} in Appendix A.

\subsection{Center of mass}
In physics, the center of mass of a distribution of mass in space is the unique point where the weighted relative position of the distributed mass sums to zero. \footnote{ \url{https://en.wikipedia.org/wiki/Center_of_mass}}

Suppose we have a system of particles, noted $\tau$, which consists of particles $P_i,  i = 1, \dots , n$,  each with mass $m_i$ and coordinates $r_i$, the coordinate $\boldsymbol {P}_\tau$ of the center of mass of $\tau$ satisfies the condition:

\begin{equation}
\sum _{i=1}^{n}m_{i}(r_{i}-{\boldsymbol {P}_\tau})=0
\label{eq:center-of-mass1}
\end{equation}

Solving this equation for $\boldsymbol {P}_\tau$  yields:

\begin{equation*}
\boldsymbol {P}_\tau ={\frac {1}{M}}\sum _{i=1}^{n}m_{i} r _{i}
\label{eq:concept-to-topic11}
\end{equation*}

where $M=\sum _{i=1}^{n}m_{i}$  is the total mass of all of the particles.

Observing Equation ~\ref{eq:center-of-mass1}, we know Center of Mass $\boldsymbol {P}_\tau$ is the unique solution of the following optimization problem:

\begin{equation*}
\min _{x\in  \mathbb{R} } \;f_\tau \left(x \right)
\end{equation*}
where $\mathbb{R}$ is the set of real numbers, and 
\begin{equation}
f_\tau \left(x \right) = \sum _{i=1}^{n}m_{i}   {\left(    x- r_i \right)}^2
\label{mass-dis1}
\end{equation}

The proof is simple, since $f_\tau \left(x \right)$ is a strictly convex and differentiable function about $x$, set its derivative equal to zero, we get its unique optimal solution.

\begin{equation*}
\frac{df_\tau}{dx} = 2  \left[     \sum _{i=1}^{n}m_{i}  \left(    x^{*}- r_i \right)   \right] = 0
\end{equation*}

\begin{equation}
\Rightarrow  \quad \sum _{i=1}^{n}m_{i}(r_{i}-{ x^{*}})=0
\label{eq:center-of-mass2}
\end{equation}

Note the similarity between Equation ~\ref{eq:center-of-mass1} and ~\ref{eq:center-of-mass2}. Therefore,

\begin{equation}
\boldsymbol {P}_\tau = x^{*} =\operatorname*{argmin}_{x\in \mathbb{R} } \;f_\tau \left(x \right)
\label{argmin1}
\end{equation}

\subsection{Semantic center of mass (SCOM)}

Based on the notion of center of mass in physics and definitions introduced previously, it is now ready to define the Semantic Center of Mass.

\begin{definition}
Semantic Center of Mass

The SCOM of a document $\tau$ is the solution of the following optimization problem:
\begin{equation*}
\min _{x\in  {\Omega} } \;g_\tau \left(x \right)
\end{equation*}
where, $\Omega$ is a set of candidate topics in domain $\Psi$, $x$ is a topic belongs to $\Omega$, and 

\begin{equation}
g_\tau \left(x \right) = \sum _{i=1}^{n}m_{i}  d^2 {\left(x, c_i \right)}
\label{mass-dis2}
\end{equation}

where $c_i$ is a concept in document $\tau$, and $m_i$ is its term frequency. $d^2 {\left(x, c_i \right)}$ is the \textbf{squared} distance from Concept $c_i$ to Topic $x$.

That is,

\begin{equation}
\boldsymbol {P}_\tau      =\operatorname*{argmin}_{x\in  {\Omega} } \;g_\tau \left(x \right)  
\label{argmin2} 
\end{equation}
\end{definition}

Note the similarity between Equation ~\ref{argmin1} and ~\ref{argmin2}, and between Equation ~\ref{mass-dis1} and ~\ref{mass-dis2}. Also note in Equation ~\ref{mass-dis2}  the \textbf{squared} distance from a concept to a topic is used, to be consistent with the definition of Center of Mass in Equation ~\ref{mass-dis1}.

Table ~\ref{tab:appA1} in Appendix A lists the squared distances from a concept to a topic in domain U-map One, except the topic that contains all the concepts in U-map One. 

\subsubsection{Choose a candidate topic set} 

Unlike Center of Mass in physics, there is no obvious choice what the candidate topic set $\Omega$ should be. Following are some examples:
\begin{align*} 
\Omega_1   &=   \{All\ the\ topics\ of\ length\ one.\}\\
\Omega_2   &=   \{All\ the\ topics\ of\ length\ two.\}\\
\Omega_3   &=   \{All\ the\ topics\  length\  less\ than\ 10.\}\\
\Omega_4   &=   \{All\ the\ topics\ in\ a\ domain.\}
\end{align*}

We use an example document $\tau$ to show how different setting of $\Omega$ affects the discovered SCOM.
\begin{align*} 
\tau   = \{  A:10, \quad B:0,\quad C:1,\quad D:5  ,\quad E:0 ,\quad F:3 \}    
\end{align*}
Document $\tau$ is a 4-concept-document with concept-length 4, and length 19 (10 + 1 + 5 + 3 = 19). Its Ingredient Topic $T_\tau$ is $ \{A,C,D,F\}$ .

By checking Table ~\ref{tab:appA1} in Appendix A:
\begin{align*} 
If\ \Omega = \Omega_1 ,  \ then\ \boldsymbol {P}_\tau =  \{C\} ;  
\end{align*}

Surprisingly, the discovered ``topic'', or SCOM, of Document $\tau$, is not the dominant concept (that is Concept A) in $\tau$. The reason is that the result is not only decided by the content of $\tau$, but also supervised by an Understanding Map.
\begin{align*} 
If\ \Omega = \Omega_2 ,  \ then\ \boldsymbol {P}_\tau =  \{A,D\} ;  
\end{align*}

This time the discovered SCOM is agree with the dominant concepts in $\tau$. That implies the content of $\tau$ does play a role to the result.

If the  candidate topic set $\Omega$ contains topics of different length, such as $\Omega_3$ and $\Omega_4$, there is a problem need some consideration. Due to our definition of concept-to-topic distance in Equation ~\ref{eq:concept-to-topic}, a concept's distance to a length-$\left(n+1\right)$ topic is systematically shorter than a length-$n$ topic. Table ~\ref{tab:staumapone} lists some statistics of U-map One. The first column ``Topic\_len'' is topic length; the second column ``Avg\_dis'' is the average squared concept-to-topic distance of corresponding group, readers can check Table ~\ref{tab:appA1} in Appendix A for the values. To compare the topics fairly, we need to normalize the squared distances, such that the expectation of a length-$\left(n+1\right)$ topic is equal to a length-$n$ topic.

\begin{table*} 
	\begin{center}
		\scalebox{0.95}{
			\begin{tabular}{rrrrrrrrr}
				\toprule
				Topic\_len &  Avg\_dis &  Norm\_factor &  Avg\_dis\_norm &  Data\_sparse &  Avg\_dis\_noise &  Norm\_factor\_noise &  Avg\_dis\_norm\_noise &  Data\_sparse\_noise \\
				\midrule
				1 &     3.83 &         1.00 &          3.83 &         0.17 &           4.03 &               1.00 &                4.03 &                0.0 \\
				2 &     1.56 &         2.46 &          3.83 &         0.33 &           1.76 &               2.30 &                4.03 &                0.0 \\
				3 &     0.79 &         4.84 &          3.83 &         0.50 &           0.99 &               4.07 &                4.03 &                0.0 \\
				4 &     0.40 &         9.58 &          3.83 &         0.67 &           0.60 &               6.72 &                4.03 &                0.0 \\
				5 &     0.17 &        23.00 &          3.83 &         0.83 &           0.37 &              11.00 &                4.03 &                0.0 \\
				\bottomrule
			\end{tabular}			
			
		}
		\caption{Some statistics of U-map One}
		\label{tab:staumapone}
	\end{center}
	
\end{table*}	

\subsubsection{Revision one of objective function} 
\begin{equation}
g_\tau \left(x \right) = \phi_x \sum _{i=1}^{n}m_{i}  d^2 {\left(x, c_i \right)} 
\label{mass-dis-fix1}
\end{equation}
We introduce a normalization factor $\phi_x$ to each length of topics, e.g., the third column of Table ~\ref{tab:staumapone}. After the normalization, the expectation of a concept to different length of topics is equal. See the fourth column ``Avg\_dis\_norm'' of Table ~\ref{tab:staumapone}. Readers can check Table ~\ref{tab:appA2} in Appendix A for the values.
\subsubsection{Revision two of objective function} 
\begin{equation}
g_\tau \left(x \right) = \frac{\phi_x}{M}  \sum _{i=1}^{n}m_{i}  d^2 {\left(x, c_i \right)} 
\label{mass-dis-fix11}
\end{equation}
where $M=\sum _{i=1}^{n}m_{i}$  is the length of Document $\tau$.
In this revision, the objective function is divided  by the length of document. Note dividing by a positive constant does not affect the optimal solution of Equation \ref{argmin2}. This revision is also the definition of a document's normalized distance to a topic.

\begin{definition}
Document-to-topic distance

Document $\tau$'s  distance to topic $x$ is:

\begin{equation}
d(\tau, x) = \frac{\phi_x}{M} \sum _{i=1}^{n}m_{i}  d^2 {\left(x, c_i \right)}   
\label{eq:document-to-topic-distance}
\end{equation}
\end{definition}

With the normalization of Revision One and Two, we can compare a document's distance to different topics, and  different documents' distances to a topic.

With this definition, Equation \ref{argmin2}  is equivalent to: 
\begin{equation}
\boldsymbol {P}_\tau      =\operatorname*{argmin}_{x\in  {\Omega} } \;d(\tau, x) 
\label{eq:argmin-d-to-t-distance} 
\end{equation}

That is, a document's SCOM is the topic that has the shortest document-to-topic distance to the document.

\begin{definition}
Document-to-domain distance

Now we can vectorize Document $\tau$ as its distances to different topics in Domain $\Psi$.

Document $\tau$'s  distance to Domain $\Psi$ is:

\begin{equation}
d(\tau, \Psi) = \{d(\tau, x) \mid x \in \Psi\}
\label{eq:document-to-domain-distance}
\end{equation}
\end{definition}

\subsection{Partiality for long topics}
\begin{figure}
	\centering
	\includegraphics[width=0.9\columnwidth]{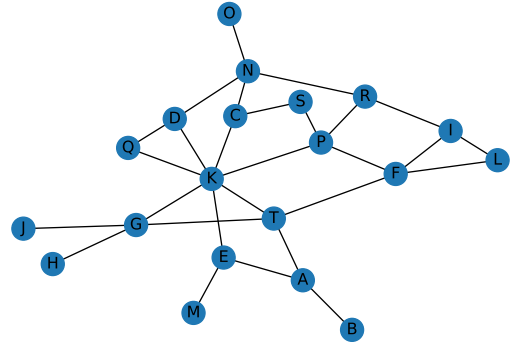}
	\caption{U-map Shallow,  the longest concept-to-concept distance is 6}
	\label{fig:shallow-umap}
\end{figure}

\begin{figure}
	\centering
	\includegraphics[width=0.9\columnwidth]{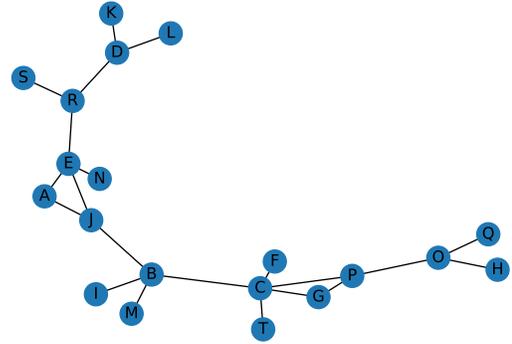}
	\caption{U-map Deep,  the longest concept-to-concept distance is 9}
	\label{fig:deep-umap}
\end{figure}

Although in Equation \ref{mass-dis-fix11} , factor $\phi_x$ is used to adjust the squared distances, such that the expectation of a document's distance to different topics are equal. There still exists some preference for long topics.
We use an experiment to show this partiality problem. In the experiment, 
we calculate three random generated documents' (Doc1, Doc2, and Doc3 in Table \ref{tab:six-doc} ) distances to two artificial Understanding Map (see Figure \ref{fig:shallow-umap} and \ref{fig:deep-umap}). 

\begin{table*} 
	\begin{center}
		\scalebox{0.95}{

\begin{tabular}{lrrrrrrrrrrrrrrrrrrrr}
	\toprule
	{} &  A &  B &    C &   D &    E &  F &   G &     H &  I &    J &   K &    L &   M &  N &    O &    P &  Q &    R &   S &   T \\
	\midrule
	Doc1 &  0 &  0 &   12 &   9 &   13 &  1 &   0 &     5 &  0 &    0 &   3 &   22 &   3 &  0 &    3 &    0 &  9 &   10 &  23 &   0 \\
	Doc2 &  1 &  3 &    2 &   0 &    0 &  0 &   0 &     9 &  3 &    1 &   0 &    2 &   0 &  0 &    0 &    0 &  1 &    0 &   0 &   0 \\
	Doc3 &  0 &  3 &    4 &  37 &    0 &  8 &  25 &     1 &  2 &    2 &   0 &   24 &  12 &  0 &   26 &   14 &  8 &    1 &  32 &  17 \\
	Doc4 &  0 &  0 &    4 &   0 &    0 &  2 &   3 &  1000 &  1 &    0 &   2 &    1 &   0 &  2 &    0 &    0 &  2 &    1 &   0 &   0 \\
	Doc5 &  0 &  0 &  100 &   0 &  100 &  0 &   0 &   100 &  0 &  100 &   0 &  100 &   0 &  0 &  100 &  100 &  0 &  100 &   0 &   0 \\
	Doc6 &  0 &  0 &    0 &   3 &    0 &  0 &   0 &     0 &  0 &    0 &  32 &   16 &   0 &  0 &    0 &    0 &  0 &    9 &   2 &   0 \\
	\bottomrule
\end{tabular}					
			
		}
		\caption{Six artificial documents represented as bag-of-concepts}
		\label{tab:six-doc}
	\end{center}
\end{table*} 

Both U-maps contain the same set of 20 concepts, but with different structures. In U-map Shallow,  the longest concept-to-concept distance is 6; for U-map Deep,  the longest concept-to-concept distance is 9. In practice, an U-map is constructed based on concepts' definitions. A concept's definition is supposed to be fixed. Therefore, there is only one U-map for supervising the interpretation of documents in practice. Here we use two U-maps with different structures to test how the analysis results are affected by U-maps.

\begin{figure} 
	\begin{subfigure}{0.45\columnwidth}
	\includegraphics[width=\linewidth]{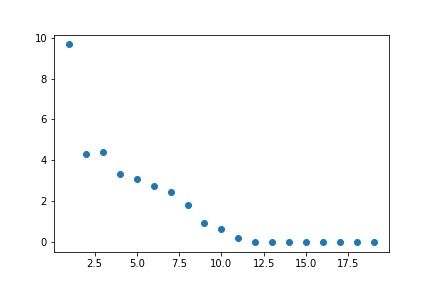}
	\caption{Doc1, U-map Deep}  
\end{subfigure}    
\begin{subfigure}{0.45\columnwidth}
	\includegraphics[width=\linewidth]{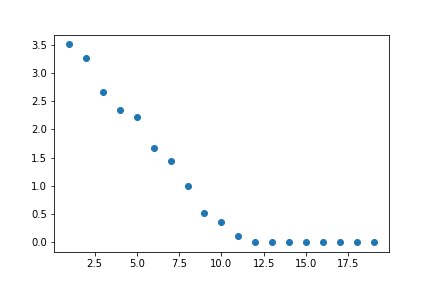}
	\caption{Doc1, U-map Shallow}  
\end{subfigure}    

\begin{subfigure}{0.45\columnwidth}
	\includegraphics[width=\linewidth]{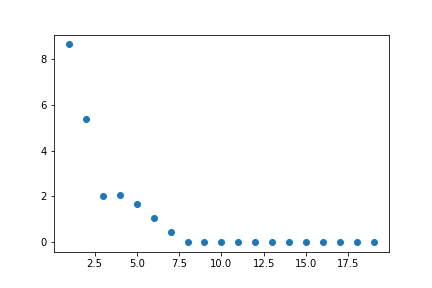}
	\caption{Doc2, U-map Deep}  
\end{subfigure}
\begin{subfigure}{0.45\columnwidth}
	\includegraphics[width=\linewidth]{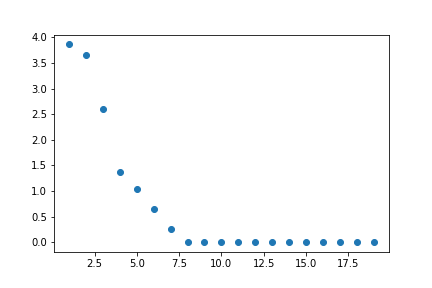}
	\caption{Doc2, U-map Shallow} \label{fig:a}
\end{subfigure}  

\begin{subfigure}{0.45\columnwidth}
	\includegraphics[width=\linewidth]{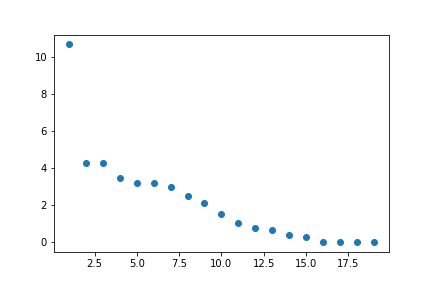}
	\caption{Doc3, U-map Deep} \label{fig:b}
\end{subfigure}    
\begin{subfigure}{0.45\columnwidth}
	\includegraphics[width=\linewidth]{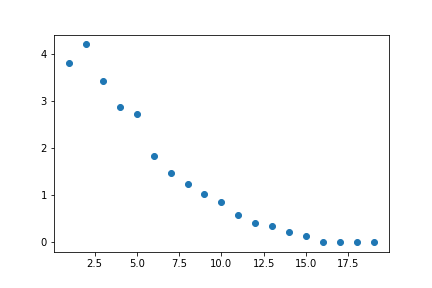}
	\caption{Doc3, U-map Shallow} \label{fig:d}
\end{subfigure}

\caption{Partiality for long topics} \label{fig:six-little}
\end{figure}

We first group topics by their length, then find out the champion of each group according to Equation \ref{eq:document-to-topic-distance} and \ref{eq:argmin-d-to-t-distance}, recording each champion's score and its topic length, then plot it. Figure \ref{fig:six-little} shows the results. $X$ axis is topic length, $Y$ axis is each champion's score. It clearly shows that for all documents and U-maps, a long topic's distance to a document is systematically  smaller than a short topic. That means if we directly compare long topics and short topics in Equation \ref{eq:argmin-d-to-t-distance}, the winner SCOM is always a long topic.  That is like comparing the capability of several champion weightlifters who belong to different weight classes. If we compare them directly without considering their weight classes, the weightlifter belonging to the largest weight class may always win.

Following we propose three possible methods to overcome this prejudice and find out the SCOM. In the setting, we assume  the  candidate topic set $\Omega$ is all the topics in a domain. In Section \ref{experi}, an experiment is conducted to check the outcome of the methods for searching the SCOM.

\subsection{Method 1:  curve fitting} \label{sec-curve-fitting}
\begin{figure}
	\centering
	\includegraphics[width=0.9\columnwidth]{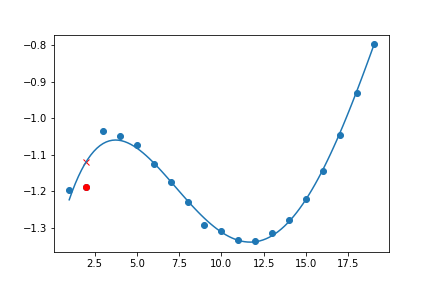}
	\caption{Curve fitting,  red cross: supposed position, red dot: actual position}
	\label{fig:three-steps1}
\end{figure}

In curve fitting method, three steps are needed to find out the SCOM. We use Doc1 in Table \ref{tab:six-doc} supervised by U-map Deep to illustrate the logic.
\begin{itemize}
	\item Step 1: Find out the  local champion of each length of topics and their distances to Doc1.
	That is $x_n^\star$ and $d(\tau, x_n^\star) $ in Equation \ref{equ:three-step1}, and Column ``Topic'' and ``Distance'' in Table \ref{tab:3-steps-table}.
	
	\item Step 2: Calculate the  mean and standard deviation (SD) of each group, and calculate each local champion's relative position in their group with Equation \ref{equ:three-step1}. That is $z(x_n^\star)$  in Equation \ref{equ:three-step1}, and Column ``Actual'' in Table \ref{tab:3-steps-table}.
	
	\item Step 3: Check if there exists some pattern (curve) between topic length and the calculated relative position. If it is, fit a curve $h(x)$, like the one in Figure \ref{fig:three-steps1}. Then calculate each local champion's supposed position on the curve. Then compare each local champion's actual position and its supposed position. Select the one that locates lowest than its supposed position (Equation \ref{equ:three-step3}). See the red dot in Figure \ref{fig:three-steps1}, that is Topic $\{P,R\}$ in Table \ref{tab:3-steps-table}.
\end{itemize}

\begin{equation}
z(x_n^\star) = \frac {d(\tau, x_n^\star)  -\mu_n }{\sigma_n }
\label{equ:three-step1}
\end{equation}

\begin{equation}
\boldsymbol {P}_\tau      =\operatorname*{argmin}_{x_n^\star \in  {\Omega^\star} } \; z(x_n^\star) -  h(x_n^\star )
\label{equ:three-step3} 
\end{equation}

\begin{table*} 
	\begin{center}
		\scalebox{0.85}{
\begin{tabular}{rlrrrrrr}
	\toprule
	Topic\_len &               Topic &    Distance &        Mean &       SD &      Actual &  Supposed &    Final \\
	\midrule
	1 &                    E &  9.672566 &  22.175221 &  10.454785 & -1.195879 & -1.223085 &  0.02721 \\
	2 &                  \bf{PR}  &  4.318913 &  22.227630 &  15.078249 & -1.187719 & -1.119204 & \bf{-0.06851}  \\
	3 &                  LPR &  4.422448 &  21.918055 &  16.905369 & -1.034914 & -1.069265 &  0.03435 \\
	4 &                 CLOR &  3.336689 &  21.475281 &  17.280260 & -1.049671 & -1.060600 &  0.01093 \\
	5 &                CELOS &  3.094624 &  21.071216 &  16.747912 & -1.073363 & -1.081698 &  0.00833 \\
	6 &               CDELOS &  2.758609 &  20.754817 &  16.004105 & -1.124475 & -1.122199 & -0.00228 \\
	7 &              CDELMOS &  2.435302 &  20.508720 &  15.404604 & -1.173248 & -1.172900 & -0.00035 \\
	8 &             CDEHLMQS &  1.835602 &  20.302052 &  15.033593 & -1.228346 & -1.225750 & -0.00260 \\
	9 &            CDEHLMQRS &  0.929391 &  20.112189 &  14.838634 & -1.292760 & -1.273853 & -0.01891 \\
	10 &           CDEHLMOQRS &  0.649870 &  19.928507 &  14.733530 & -1.308487 & -1.311466 &  0.00298 \\
	11 &          CDEHKLMOQRS &  0.198685 &  19.749711 &  14.668950 & -1.332817 & -1.334002 &  0.00119 \\
	12 &         CDEFHKLMOQRS &  0.000000 &  19.579258 &  14.659502 & -1.335602 & -1.338027 &  0.00243 \\
	13 &        CDEFHKLMNOQRS &  0.000000 &  19.421062 &  14.769382 & -1.314954 & -1.321261 &  0.00631 \\
	14 &       BCDEFHKLMOPQRS &  0.000000 &  19.276552 &  15.080723 & -1.278225 & -1.282579 &  0.00435 \\
	15 &      ACDEFGHKLMNOQRS &  0.000000 &  19.143185 &  15.673610 & -1.221364 & -1.222009 &  0.00065 \\
	16 &     BCDEFGHKLMNOPQRS &  0.000000 &  19.013675 &  16.630109 & -1.143328 & -1.140734 & -0.00259 \\
	17 &    BCDEFGHKLMNOPQRST &  0.000000 &  18.875118 &  18.057134 & -1.045300 & -1.041091 & -0.00421 \\
	18 &   CDEFGHIJKLMNOPQRST &  0.000000 &  18.707360 &  20.123152 & -0.929644 & -0.926570 & -0.00307 \\
	19 &  ABCDEFGHIJKLMNOPQRS &  0.000000 &  18.480000 &  23.157493 & -0.798014 & -0.801816 &  0.00380 \\
	\bottomrule
\end{tabular}

		}
		\caption{Doc1 supervised by U-map Deep, Method 1}
		\label{tab:3-steps-table}
	\end{center}
	
\end{table*}

Therefore, according to Method 1, Doc1  supervised by U-map Deep, the final SCOM is Topic $\{P,R\}$. Note in this method the normalization factor $\phi_x$ in Equation \ref{eq:document-to-topic-distance}  is not necessary.  Since Equation \ref{equ:three-step1} has neutralized its effect.

\subsection{Method 2: AIC/BIC-like penalty} 

In model selection, it is possible to increase the likelihood by adding parameters, but doing so may result in over-fitting. Both Akaike information criterion (AIC) and Bayesian information criterion (BIC)  attempt to resolve this problem by introducing a penalty term for the number of parameters in the model \cite{sakamoto1986akaike, schwarz1978estimating}.  Here we are facing a similar problem like in model selection, hence similar penalty term like in AIC/BIC can be used.
\subsubsection{Revision three of objective function} 
\begin{equation}
d(\tau, x)  =   \alpha L_x  + \frac{\phi_x}{M} \sum _{i=1}^{n}m_{i}  d^2 {\left(x, c_i \right)}  
\label{mass-dis-fix32}
\end{equation}

\begin{figure} 
	\begin{subfigure}{0.45\columnwidth}
	\includegraphics[width=\linewidth]{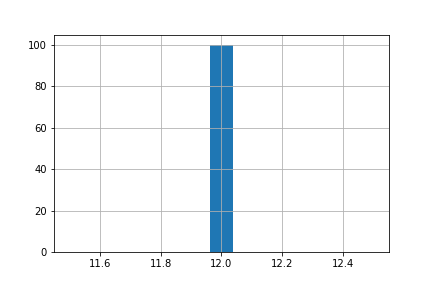}
	\caption{$\alpha$ = 0.1}  
\end{subfigure}    
\begin{subfigure}{0.45\columnwidth}
	\includegraphics[width=\linewidth]{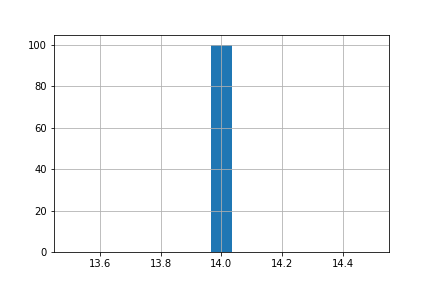}
	\caption{$\delta$ = 0.01}  
\end{subfigure}    

	\begin{subfigure}{0.45\columnwidth}
	\includegraphics[width=\linewidth]{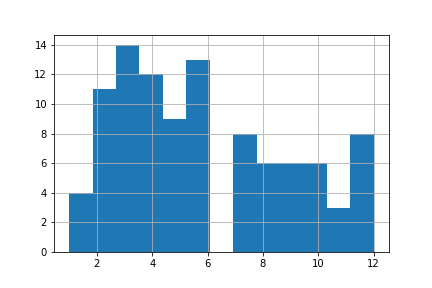}
	\caption{$\alpha$ = 0.4}  
\end{subfigure}    
\begin{subfigure}{0.45\columnwidth}
	\includegraphics[width=\linewidth]{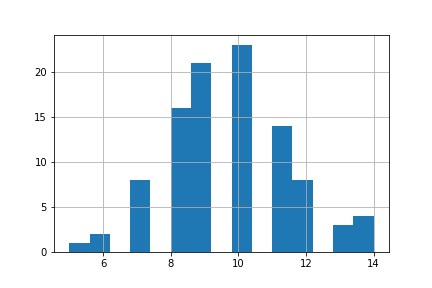}
	\caption{$\delta$ = 0.1}  
\end{subfigure}   

	\begin{subfigure}{0.45\columnwidth}
	\includegraphics[width=\linewidth]{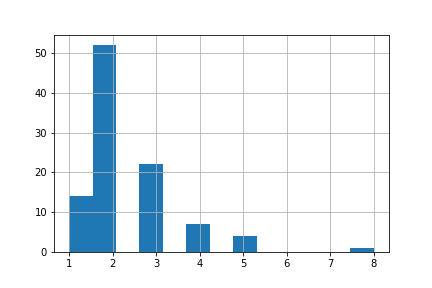}
	\caption{$\alpha$ = 1}  
\end{subfigure}       
\begin{subfigure}{0.45\columnwidth}
	\includegraphics[width=\linewidth]{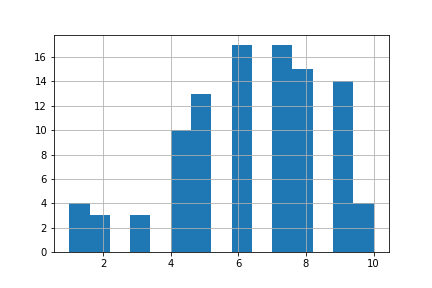}
	\caption{$\delta$ = 0.3}  
\end{subfigure}  

	\begin{subfigure}{0.45\columnwidth}
	\includegraphics[width=\linewidth]{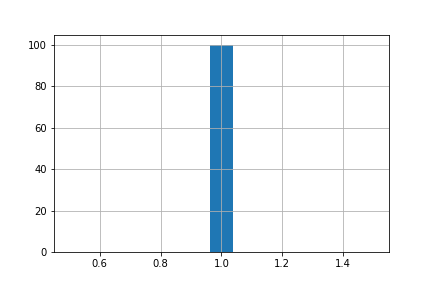}
	\caption{$\alpha$ = 10}  
\end{subfigure}     
\begin{subfigure}{0.45\columnwidth}
	\includegraphics[width=\linewidth]{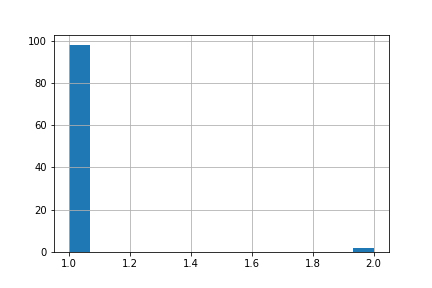}
	\caption{$\delta$ = 100}  
\end{subfigure}   

\caption{Effects of parameter $\alpha$ and $\delta$} \label{fig:8-little}
\end{figure}

Where $L_x$ is topic $x$'s length, $\alpha$ is a parameter that adjusting the length of the discovered SCOM. An experiment is conducted to test how the  $\alpha$ parameter may affect the results of Equation \ref{eq:argmin-d-to-t-distance}. In the experiment, we randomly generated 100 documents, with their concept-length fixed to 12, and varying document length. Supervised by U-map Deep, then use Equation \ref{mass-dis-fix32} and \ref{eq:argmin-d-to-t-distance} for searching the SCOM. The result is showed in the left column of Figure \ref{fig:8-little}. 

It can be seen that when $\alpha$ is small (0.1), all the discovered SCOM has length 12, that is the documents' concept-length. When 
$\alpha$ is large enough (e.g., 10), all the discovered SCOM converge to length 1. When $\alpha$ is moderate (e.g., 0.4), other length of topics then get the chance to win the competition. By setting $\alpha$ to be an arithmetic progression until the discovered SCOM converges to length one. We can count how many times each local champion wins, and set the SCOM as the local champion which has the largest counting votes.  Table \ref{aic-doc1} in Appendix A is an example when Doc1 in Table \ref{tab:six-doc} is supervised by U-map Deep. It can be seen Topic $\{P,R\}$ gets the most votes, so it is the discovered SCOM. Interestingly, the winner is stable to the common difference of the arithmetic sequence. When the SCOM is a length one topic, the evidence is that the sequence converges to length one when $\alpha$ is a small number (e.g., 0.26).

\subsection{Method 3:  introduce some noise}
The speculated reason for the preference of long topics is data sparsity. E.g., the fifth column ``Data\_sparse''  of Table \ref{tab:staumapone} calculates the ratio of data sparsity in each length group of topics. E.g., 0.67 means 67\% of the squared distances are zero (check Table \ref{tab:appA1} in Appendix A). To deal with data sparsity, some noise is introduced to the squared distances (see Table \ref{tab:appA3} in Appendix A). After introduction of some noise, the data sparsity problem is solved, see the ninth column ``Data\_sparse\_noise''  of Table \ref{tab:staumapone} and Table \ref{tab:appA4} in Appendix A. The following objective function is used in Method 3.
\subsubsection{Revision four of objective function} 
\begin{equation}
d(\tau, x)  = \frac{ \phi_x(\delta )}{M} \sum _{i=1}^{n}m_{i}    \left[   d^2 {\left(x, c_i \right)}  + \delta       \right]          
\label{mass-dis-fix4}
\end{equation}
where $\delta$ is the noise parameter. Note in this revision, the normalization factor $\phi_x$ is a function of the noise parameter $\delta$; different $\delta$ results in different $\phi_x$. See the ``Norm\_factor\_noise'' column of Table \ref{tab:staumapone}, the values are calculated when $\delta$ are set to be $0.2$.  Readers can check Table \ref{tab:appA3} for the values.

Like in Method 2, an experiment is conducted to test how the  $\delta$ parameter may affect the results of Equation \ref{eq:argmin-d-to-t-distance}. In the experiment, we randomly generated 100 documents, with their concept-length fixed to 14, and varying document length. Supervised by U-map Shallow, then use Equation \ref{mass-dis-fix4} and \ref{eq:argmin-d-to-t-distance} for searching the SCOM. The result is showed in the right column of Figure \ref{fig:8-little}.  

It can be seen that when $\delta$ is small (0.01), all the discovered SCOM has length 14, that is the documents' concept-length. When $\delta$ is large enough (e.g., 100), almost all of the discovered SCOM converge to length 1, with some little exception converges to length 2. When $\delta$ is moderate (e.g., 0.3), other length of topics get the chance of winning the competition. 

By setting $\delta$ to be an arithmetic progression until the discovered SCOM converges to length one. Like in Method 2, we can count how many times each local champion wins, and set the SCOM as the local champion which has the largest counting votes.  Table \ref{noise-doc1} in Appendix A is an example when Doc1 in Table \ref{tab:six-doc} is supervised by U-map Deep. It can be seen Topic $\{P,R\}$ gets the most votes, so it is the discovered SCOM. Again, the winner is stable to the common difference of the arithmetic sequence.

Comparing the three methods, it seems Method 1 is theoretically more reasonable.

\begin{table}
	\begin{center}
		\scalebox{0.7}{

\begin{tabular}{lllllll}
	\toprule
	{} & Doc1      & Doc2 & Doc3    & Doc4 & Doc5     & Doc6 \\
	\midrule
	deep-noise             & PR        & BHL  & CD      & H    & PR/OR    & KLR  \\
	deep-AIC               & PR        & BHL  & CD      & H    & PR/OR    & KLR  \\
	deep-curve\_fitting    & PR        & BHL  & CD      & H    & CEHJLOPR & KLR  \\
	shallow-noise          & CIK       & BHIK & DEGLOS  & H    & CEHJLOPR & IK   \\
	shallow-AIC            & CDEHLOQRS & BHIK & DGLMOST & H    & CEHJLOPR & IK   \\
	shallow-curve\_fitting & CDEHLOQRS & BHIK & DEGLOS  & H    & CEHJLOPR & IK \\
	\bottomrule 
\end{tabular}

			}
\caption{Discovered SCOM of six documents}
\label{tab:scom}
	\end{center}
\end{table}

\section{More experiments}  \label{experi}
In this section, more experiments are conducted to test the outcome of UM-S-TM.
\subsection{Comparing the three methods} 
In this experiment, six artificial documents of Table \ref{tab:six-doc} are analyzed with the three methods, supervised by U-map Deep and Shallow respectively. The first three documents are randomly generated; the last three are manually generated, to test how UM-S-TM behaves under some extreme conditions. E.g., Doc4 has a dominant concept that has a very large term frequency than other concepts; in Doc5 all the eight concepts have the same term frequency; in Doc6 all the five concepts sit on a corner of U-map Deep.

\begin{figure*} 
\begin{subfigure}{0.16\textwidth}
\includegraphics[width=\linewidth]{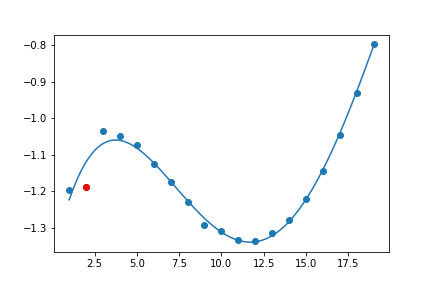}
\caption{Doc1, Deep}  
\end{subfigure}    
\begin{subfigure}{0.16\textwidth}
	\includegraphics[width=\linewidth]{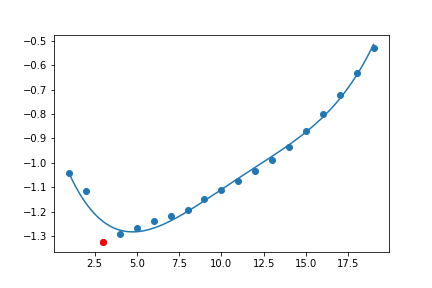}
	\caption{Doc2, Deep}  
\end{subfigure}
\begin{subfigure}{0.16\textwidth}
	\includegraphics[width=\linewidth]{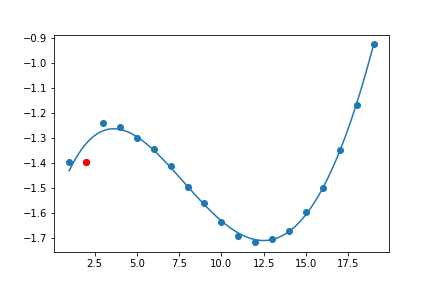}
	\caption{Doc3, Deep}  
\end{subfigure}    
\begin{subfigure}{0.16\textwidth}
	\includegraphics[width=\linewidth]{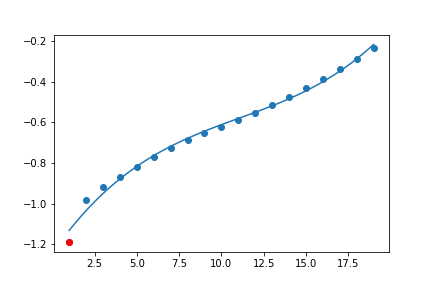}
	\caption{Doc4, Deep}  
\end{subfigure}    
\begin{subfigure}{0.16\textwidth}
	\includegraphics[width=\linewidth]{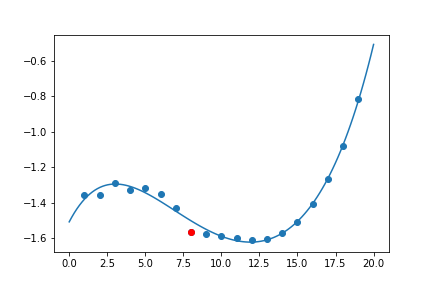}
	\caption{Doc5, Deep}  
\end{subfigure}    
\begin{subfigure}{0.16\textwidth}
	\includegraphics[width=\linewidth]{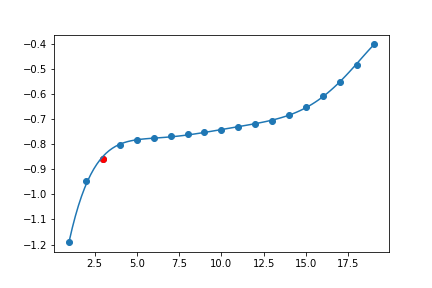}
	\caption{Doc6, Deep}  
\end{subfigure}         

\begin{subfigure}{0.16\textwidth}
	\includegraphics[width=\linewidth]{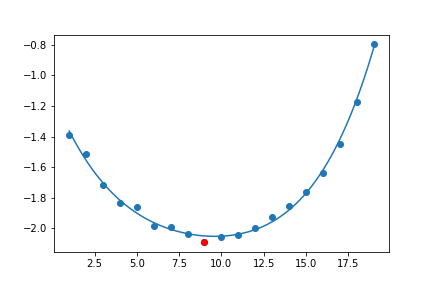}
	\caption{Doc1, Shallow}  
\end{subfigure}    
\begin{subfigure}{0.16\textwidth}
	\includegraphics[width=\linewidth]{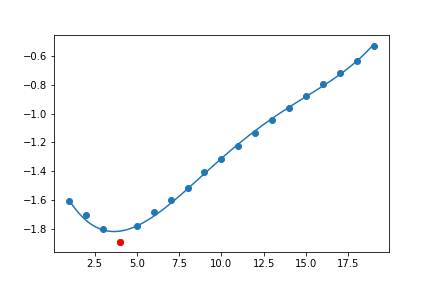}
	\caption{Doc2, Shallow}  
\end{subfigure}
\begin{subfigure}{0.16\textwidth}
	\includegraphics[width=\linewidth]{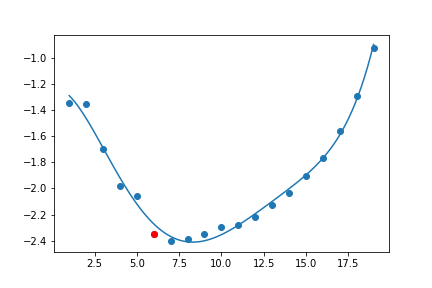}
	\caption{Doc3, Shallow}  
\end{subfigure}    
\begin{subfigure}{0.16\textwidth}
	\includegraphics[width=\linewidth]{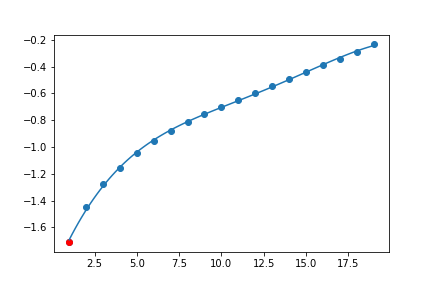}
	\caption{Doc4, Shallow}  
\end{subfigure}    
\begin{subfigure}{0.16\textwidth}
	\includegraphics[width=\linewidth]{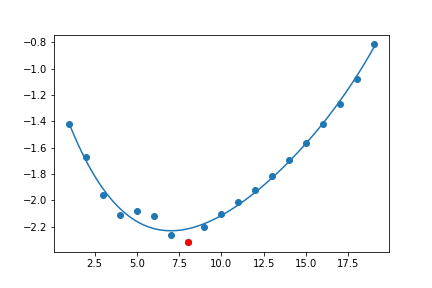}
	\caption{Doc5, Shallow}  
\end{subfigure}    
\begin{subfigure}{0.16\textwidth}
	\includegraphics[width=\linewidth]{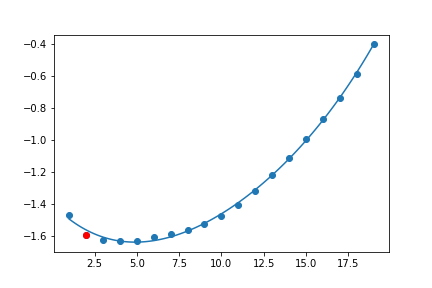}
	\caption{Doc6, Shallow}  
\end{subfigure}

\caption{Discovered SCOM of six documents, Method 1} 
\label{fig:12-little}
\end{figure*}

Table \ref{tab:scom} shows the results. Figure \ref{fig:12-little} shows the fitted curves and the SCOMs selected by Method 1. It can be seen that most of time the three methods are agreeing with each other. In three situations - Doc1 supervised by U-map Shallow, Doc3 supervised by U-map Shallow, Doc5 supervised by U-map Deep - there are some disputes.

By checking the experiment data, even there are disputes, the results are not far from each other. One method's champion is usually another method's second place or third place. 

Table \ref{tab:agreement} lists 22 more test cases. The 22 documents are randomly generated with varying concept-length and document length. If the three methods have a unanimous agreement, then all of them are scored one; if one of them is disagreeing with the other two, then the two are scored one, the disagreeing one is scored zero; if all of them have different opinion, means there is not any agreement, then all of them are scored zero.

Summarizing all the 34 test cases of Table \ref{tab:scom} and \ref{tab:agreement}, 68\% of the cases, there are unanimous agreements. 97\% of the cases, at least two methods are agreeing with each other about the SCOM. There is only one case, they have totally different opinion. Further study is necessary to check efficacy of the three methods.

\begin{table}[]
	
	\scalebox{0.8}{
\begin{tabular}{@{}llllll@{}}
\toprule
Doc\_ID & Curve\_fitting & AIC & Noise & U-map   & Agreement \\ \midrule
1       & 1              & 1   & 1     & Shallow & 3         \\
2       & 1              & 1   & 1     & Shallow & 3         \\
3       & 1              & 1   & 1     & Shallow & 3         \\
4       & 1              & 1   & 1     & Shallow & 3         \\
5       & 1              & 1   & 1     & Shallow & 3         \\
6       & 0              & 1   & 1     & Shallow & 2         \\
7       & 1              & 1   & 0     & Shallow & 2         \\
8       & 1              & 1   & 1     & Shallow & 3         \\
9       & 1              & 1   & 1     & Shallow & 3         \\
10      & 1              & 0   & 1     & Shallow & 2         \\
11      & 1              & 1   & 1     & Shallow & 3         \\
12      & 1              & 1   & 0     & Deep    & 2         \\
13      & 0              & 1   & 1     & Deep    & 2         \\
14      & 0              & 1   & 1     & Deep    & 2         \\
15      & 0              & 1   & 1     & Deep    & 2         \\
16      & 0              & 0   & 0     & Deep    & 0         \\
17      & 0              & 1   & 1     & Deep    & 2         \\
18      & 1              & 1   & 1     & Deep    & 3         \\
19      & 1              & 1   & 1     & Deep    & 3         \\
20      & 1              & 1   & 1     & Deep    & 3         \\
21      & 1              & 1   & 1     & Deep    & 3         \\
22      & 1              & 1   & 0     & Deep    & 2         \\ \bottomrule
\end{tabular} }
\caption{Agreement of the three methods}
\label{tab:agreement}
\end{table}

\subsection{Vectorization of documents} 
Equation \ref{eq:document-to-domain-distance} defines a method to vectorize a document. That is the document's distances to different topics of the domain. In this experiment, the vectorization is
compared with term frequency vectorization. We randomly generated 300 documents and set Doc0 as the target, finding the top10 documents that have high cosine similarity with Doc0, based on different vectorization methods.

The second row of Table \ref{simi-top10} is Doc0, the following ten rows are the top10 documents that have high cosine similarity with Doc0, measured by term frequency. The right most column is their ``Doc\_ID''.  Table \ref{300-simi-rank} shows the ``Doc\_ID'' of top10 documents based on UM-S-TM vectorization, compared with term frequency vectorization. It can be seen that supervised by different U-maps, the selected documents and their ranks are different. ``Deep-no-norm'' means U-map Deep is used without the normalization factor $\phi_x$.

\begin{table}
	\begin{center}
		\scalebox{0.7}{
			
\begin{tabular}{@{}llllll@{}}
	\toprule
	Rank & TF  & Deep & Deep-no-norm & Shallow & Shallow-no-norm \\ \midrule
	1    & 142 & 142  & 142          & 282     & 266             \\
	2    & 279 & 279  & 279          & 142     & 282             \\
	3    & 282 & 257  & 257          & 266     & 142             \\
	4    & 278 & 278  & 278          & 257     & 104             \\
	5    & 257 & 232  & 232          & 104     & 212             \\
	6    & 172 & 188  & 112          & 279     & 257             \\
	7    & 121 & 282  & 282          & 212     & 269             \\
	8    & 8   & 112  & 188          & 121     & 248             \\
	9    & 112 & 172  & 172          & 278     & 121             \\
	10   & 212 & 121  & 8            & 248     & 279             \\ \bottomrule
\end{tabular}
			
		}
		\caption{Top 10 documents similar to Doc0}
		\label{300-simi-rank}
	\end{center}
\end{table}

\begin{table}
	\begin{center}
		\scalebox{0.6}{
			
\begin{tabular}{lrrrrrrrrrrrrrrrrrrrrr}
	\toprule
	Rank &   A &  B &   C &  D &   E &  F &  G &  H &  I &  J &  K &  L &   M &  N &  O &  P &  Q &   R &  S &   T &  Doc\_ID \\
	\midrule
	   &   0 &  0 &   9 &  0 &   0 &  0 &  0 &  0 &  0 &  0 &  0 &  0 &   0 &  0 &  1 &  0 &  0 &  12 &  0 &  24 &       0 \\
	1  &   1 &  5 &  13 &  0 &   5 &  1 &  0 &  0 &  0 &  0 &  0 &  0 &   0 &  0 &  0 &  1 &  0 &   0 &  0 &  27 &     142 \\
	2  &   0 &  0 &   0 &  2 &   0 &  1 &  3 &  0 &  0 &  0 &  0 &  0 &   0 &  0 &  0 &  0 &  0 &   0 &  0 &  27 &     279 \\
	3  &   0 &  0 &   0 &  0 &   0 &  1 &  0 &  0 &  0 &  0 &  0 &  1 &   0 &  0 &  0 &  0 &  0 &   6 &  0 &   4 &     282 \\
	4  &   1 &  1 &   0 &  3 &   0 &  0 &  0 &  0 &  0 &  0 &  0 &  8 &   0 &  0 &  0 &  0 &  2 &   0 &  0 &  24 &     278 \\
	5  &   0 &  0 &  23 &  0 &  16 &  0 &  0 &  0 &  0 &  0 &  0 &  0 &   0 &  0 &  0 &  0 &  2 &   0 &  0 &  27 &     257 \\
	6  &   0 &  1 &   0 &  0 &   0 &  0 &  0 &  0 &  4 &  0 &  0 &  0 &   8 &  0 &  1 &  0 &  0 &   0 &  0 &  19 &     172 \\
	7  &   0 &  0 &   0 &  0 &   0 &  0 &  0 &  0 &  0 &  0 &  0 &  0 &   1 &  0 &  0 &  0 &  0 &   6 &  0 &   3 &     121 \\
	8  &   0 &  0 &   0 &  0 &   0 &  0 &  0 &  0 &  0 &  0 &  0 &  0 &  13 &  1 &  0 &  0 &  0 &   0 &  0 &  22 &       8 \\
	9  &   0 &  0 &   0 &  0 &   7 &  0 &  3 &  0 &  1 &  0 &  0 &  0 &  11 &  0 &  0 &  0 &  0 &   0 &  0 &  22 &     112 \\
	10 &  14 &  0 &   4 &  1 &   0 &  0 &  0 &  0 &  0 &  0 &  0 &  0 &   0 &  0 &  6 &  3 &  6 &   7 &  0 &  14 &     212 \\
	\bottomrule
\end{tabular}

		}
		\caption{Top 10 documents similar to Doc0 measured by TF}
		\label{simi-top10}
	\end{center}
\end{table}

\subsection{Capturing sequential information} 
Some information of text is stored in the sequence of concepts. E.g., ``Bob loves Alice'' has a different meaning from ``Alice loves Bob''. In UM-S-TM, it is easy to capture the sequential information of text, by calculating a sequential-document's distance to a domain.
\begin{definition}
Sequential-document-to-domain distance	

Sequential-document $\tilde{\tau}$'s  distance to Domain $\Psi$ is:

\begin{equation}
d(\tilde{\tau}, \Psi) = \frac{1}{M} \sum _{i=1}^{M} d(\tau_i, \Psi)
\label{eq:seq-document-to-domain-distance}
\end{equation}
\end{definition}
$M$ is sequential-document $\tilde{\tau}$'s length, each $d(\tau_i, \Psi)$ is a vector, defined by Equation \ref{eq:document-to-domain-distance}.
The sum is an element-wise sum. Document $\tau_i$ is generated by cutting from the first concept to the $i$th concept of sequential-document $\tilde{\tau}$, then deem it as a bag-of-concepts. So there are $M$ such documents.

Following are four simple sequential-documents. We vectorize them with Equation \ref{eq:seq-document-to-domain-distance}, and set  S0 as the target, calculating the other three's cosine similarity with S0, supervised by different U-maps. Table \ref{tab:sequential} shows the result. It can be seen that sequential-document S1 and S3 have different meaning than S0, otherwise their cosine similarity should be one. That suggests UM-S-TM has captured the sequential information in text.

\begin{align*} 
S0   &= \left[   A, \quad B,\quad C       \right]      \\
S1   &=  \left[   C, \quad B,\quad A       \right]         \\
S2   &=  \left[    A, \quad B,\quad D    \right]        \\
S3   &= \left[ A, \quad C,\quad B        \right]       
\end{align*}

\begin{table}[]
\begin{tabular}{@{}llll@{}}
\toprule
        & S1   & S2   & S3   \\ \midrule
Deep    & 0.7092 & 0.991  & 0.9891 \\
Shallow & 0.8778 & 0.9948 & 0.9714 \\ \bottomrule
\end{tabular}
\caption{Cosine similarity to Document S0}
\label{tab:sequential}
\end{table}

\subsubsection{Utilizing sentence or paragraph information} 
A document usually is composed of  sentences and paragraphs. This information can be utilized for capturing sequential information. Suppose sequential-document  $\tilde{\tau}$ is segmented into $P$ sentences or paragraphs. Equation \ref{eq:seq-document-to-domain-distance-para} can be used for calculating sequential-document $\tilde{\tau}$'s  distance to Domain $\Psi$. Each sentence or paragraph $\tilde{\tau_j}$'s distance to Domain $\Psi$, $d(\tilde{\tau_j}, \Psi)$, is calculated with Equation \ref{eq:seq-document-to-domain-distance} .
\begin{equation} 
	d(\tilde{\tau}, \Psi) =\frac{1}{P} \sum _{j=1}^{P} d(\tilde{\tau_j}, \Psi) \\
	\label{eq:seq-document-to-domain-distance-para}
\end{equation}

\section{Related work} \label{sec-related-work}
Probabilistic topic models have gained great popularity in recent years \cite{hofmann1999probabilistic,blei2003latent}. UM-S-TM has some similarity to them. Following compare the two.
\begin{itemize}
\item Probabilistic topic models like Latent Dirichlet Allocation (LDA) and probabilistic Latent Semantic Analysis (pLSA) are unsupervised models. UM-S-TM is supervised, interpretation of observed data is supervised by a semantic network (U-map).

\item In probabilistic topic models, a topic is a distribution over concepts;
in UM-S-TM, a topic is a set of concepts.

\item In probabilistic topic models, a document is expressed as a distribution over a set of topics; in UM-S-TM, a document is expressed as a vector of distances to a set of topics.
	
\item In probabilistic topic models, a document can be associated with a topic by finding the dominant topic of the document; in UM-S-TM, a document can be associated with a topic by finding the SCOM of the document.	
\end{itemize}

\section{Discussion}
\subsection{A conjecture}
By observing the experiment data, the following conjecture is speculated:

\emph{When the candidate topic set $\Omega$ is all the topics in a domain, the length of a document's SCOM is always less than or equal to its concept-length.}

If it is true, it can reduce the search scope when finding the SCOM.

When the candidate topic set is  $\{All\ the\ topics\ of\ length\ one\}$, the obtained SCOM is called SCOM-of-length-one (or One-SCOM); when the candidate topic set  $\Omega$ is all the topics in a domain, the obtained SCOM is called the Global SCOM of document $\tau$. If not explicitly stated, when we say the  SCOM of a document, we mean the Global SCOM of the document.

 It is worth noting that this is not a rigorous conjecture because we do not have a rigorous definition of the Global SCOM of a document now.

\subsection{Alternative choices}
There are other choices of defining concept-to-topic distance. E.g., defining it as the average distance from a concept to all the concepts belonging to a topic. Preliminary test shows that if this definition is used, the result of discovered SCOM is dominated by the structure of U-map. The devising aim of UM-S-TM is to let both the document content and U-map play a role, in interpreting the meaning of a document, the result should not be dominated by either of them.

To be consistent with the notion of center of mass in physics, squared distance  is used in Equation \ref{mass-dis2}, an alternative choice is not squaring the distance. Preliminary test shows that it has similar effects with squared distance.

\subsection{Too many topics} \label{Too-many}
In this paper, simple artificial U-maps containing 20 concepts are used for illustrating the mechanism of UM-S-TM. This setting let us have the capability to compare all the $2^{20} - 2 = 1,048,574$ topics in the domain by enumeration. A real U-map may contain thousands or millions of concepts. For an U-map containing one million concepts, there are $2^{(1\ million)}$ possible topics, such huge a vector cannot be processed by machines at present. Following strategies can be utilized to solve this problem.

\begin{itemize}
	\item Use graph partition techniques. With graph partition, we can segment a huge domain into simple subdomains. E.g., for a one million concepts domain, we segment it into 1,000 subdomains, then there are only 1,000 topics in this domain. Therefore, UM-S-TM is flexible for both dimension reduction and dimension expansion.
	\item Only consider important and well known concepts. By removing unimportant concepts, an U-map can be simplified.
	\item Only consider topics length less than $P$. E.g., for a one million concepts domain, if we only consider topics length less than 10, then the magnitude of topics is $10^{60}$, which is processable by machines.
	\item Only consider a connected sub-graph of an U-map as a valid topic. E.g., for U-map Deep, topic $\{ K, D , L \}$ is a valid topic because it is a  connected sub-graph; topic $\{ K, D , Q \}$ is not a valid topic because it is not a  connected sub-graph. With this constraint, the quantity of topics of U-map Deep can be reduced from about one million to about ten thousands.
\end{itemize}

\section{More Discussion}
\begin{figure}
	\centering
	\includegraphics[width=0.9\columnwidth]{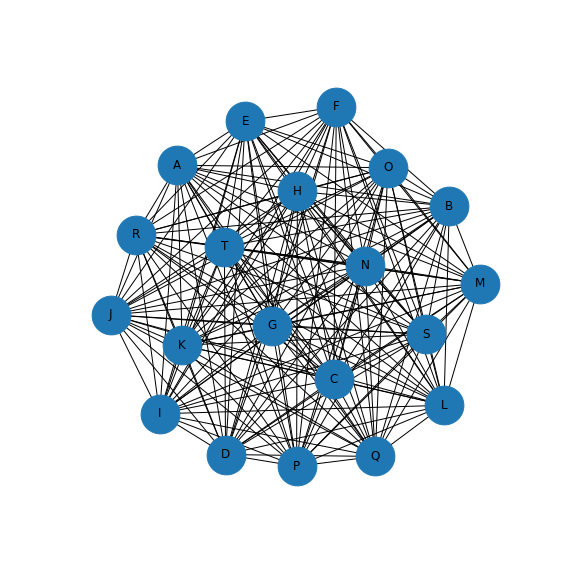}
	\caption{A complete graph of the nodes of U-map Deep and Shallow}
	\label{fig:complete-graph}
\end{figure}

In previous sections, we use UM-S-TM to analyze documents interpreted by a semantic network. Further extension may use other networks like computer networks or social networks to analyze something like documents. If we have unlimited computing power or smart methods to process huge vectors efficiently, such that we do not need to make compromises listed in Section \ref{Too-many}. Then the details of  a network is not necessary; only the squared distances between each pair of nodes are needed. That is the first six rows (seven rows if the head row is counted) of Table \ref{tab:appA1} in Appendix A. With these six rows, we can reconstruct Table \ref{tab:appA1}. That is to say, we only need to construct a complete graph like Figure \ref{fig:complete-graph}, the weight of an edge is the squared distance between the pair of nodes.

Following we propose two big questions.

\subsection{What is the SCOM of human civilization?}
To answer this question, we first make a list of human civilization spots, such as Rome, Paris, Pompeii, Maya city, Loulan Kingdom, New York, Beijing, Tokyo etc. The mass of each civilization spot is the population of it at time $t$. The distance is the natural distance between two spots. Then we construct a complete graph like Figure \ref{fig:complete-graph} and a table like Table \ref{tab:appA1}. Then, we can analyze the  SCOM of human civilization with UM-S-TM.

\subsubsection{Center of mass on a manifold}

Suppose a system of $n$ particles distributed on the surface of a ball. What is the ``center of mass'' of them on the surface? Does the following optimization problem have an analytical solution?

\begin{equation}
\boldsymbol {P}     =\operatorname*{argmin}_{c\in  {\Omega} } \;\sum _{i=1}^{n}m_{i}  d^2 {\left(c, p_i \right)}  
\end{equation}

Where $\Omega$ is the surface of the ball, $m_i$ is particle $p_i$'s mass.
$d^2 \left(c, p_i \right)$ is the squared distanced from  $c$ to particle $p_i$ on the surface. That is to say, it is the squared great-circle distance. Suppose we know each particle's coordinates.

If the optimization problem has an analytical solution, then we can conduct K-means clustering on the surface of a ball.

\subsection{What is the SCOM of the universe?}
If the number of celestial bodies in the universe is finite, then we can use UM-S-TM to analyze the SCOM of the universe. As before, we construct a complete graph like Figure \ref{fig:complete-graph} and a table like Table \ref{tab:appA1}. The mass and distance information is obvious to obtain. 

\subsection{Compared with K-means and medoid}
K-means clustering aims to partition $n$ observations into $K$ clusters in which each observation belongs to the cluster with the nearest mean (cluster centers or cluster centroid), serving as a prototype of the cluster. However, in the ``update step'' of K-means, we need the coordinates of each data point to re-compute the centroid for each cluster, which is unavailable on graph. Therefore, K-means cannot work directly on graph. SCOM and UM-S-TM have the advantage of  working directly on graphs like U-maps.

Medoids are representative objects of a data set or a cluster within a data set whose sum of dissimilarities to all the objects in the cluster is minimal. Medoids are similar in concept to means or centroids, but medoids are always restricted to be members of the data set \cite{struyf1997clustering}.  

Let ${\mathcal {X}}:=\{x_{1},x_{2},\dots ,x_{n}\}$ be a set of $n$ points in a space with distance function $d$. Medoid is defined as: \footnote{ \url{https://en.wikipedia.org/wiki/Medoid}}

\begin{equation}
	x_{\text{medoid}}=\arg \min _{y\in {\mathcal {X}}}\sum _{i=1}^{n}d(y,x_{i})
	\label{equ:medoid}
\end{equation}
Note in Equation \ref{equ:medoid} we do not require the distance being squared . Both non-squared distance and squared distance are acceptable, depending on how the dissimilarity metric between two points in a cluster is defined. However, in SCOM, only the squared distance is acceptable, otherwise the correspondence between SCOM and COM (center of mass) is lost.

Another important difference between SCOM and medoid is that ``medoids are always restricted to be members of the data set''.  SCOM does not have this restriction. E.g., in Section \ref{sec-two}, all the three methods suggest Topic $\{P,R\}$  is the SCOM of Doc1 supervised by U-map Deep. However, the term frequency of Concept `P'  in Doc1 is zero, that is to say, Concept `P' does not appear in Doc1. 

The similarity of SCOM and medoid is that both of them can work on situations where coordinates are unavailable, pair-wise distance information is sufficient.

\subsection{Approximations} \label{sec-approxi}
As discussed in Section \ref{Too-many}, when the quantity of nodes on an U-map is large, complete enumeration of possible topics in a domain is impractical. Except for the compromises listed  in Section \ref{Too-many}, several approximation methods can be employed.
\subsubsection{Giving each vertex a coordinate}   \label{coordinate}
We can allocate a coordinate to each vertex by measuring its distances to all the nodes on the graph. E.g., on U-map One, Concept C's distances to Concepts $\left[A,B,C,D,E,F\right]$ are  $\left[1,2,0,1,1,2\right]$, this vector can serve as the coordinate of Concept C. Alternative choice is the squared distance. If each data point has coordinate, then other clustering techniques like K-means and Gaussian mixture model can be used. The coordinates can also be simplified with dimension reduction techniques like Principal Component Analysis (PCA).

\subsubsection{K-means on graph}
In K-means clustering, during the ``update step'', we need the coordinates of each data point to re-compute the centroid for each cluster, which is unavailable on graph. Although the new centroid for each cluster is unavailable, but we know how to calculate the SCOM-of-length-one. It is the counterpart of center of mass or centroid on graph. So it can be used as a substitution of the new centroid. The good news is, calculation of  SCOM-of-length-one is not expensive, the complexity is bounded by $O(n^{2})$, where $n$ is the quantity of nodes on an U-map. If the ``re-computing of the new centroid'' problem is solved, then K-means clustering can work directly on a graph. Note that the ``assignment step'' of K-means clustering is easy to conduct on a graph, because we have the distance information.

Also note that in this approximation,  the approximation of Section \ref{coordinate} is not necessary. Because vectorization of vertexes of a graph may introduce extra noise. We can be exempt from this extra noise by finding  the SCOM-of-length-one by 
complete enumeration of all the possibilities. There are only $n$ possibilities, where $n$ is the number of nodes on the graph. Therefore, this optimization step is not expensive.

If we want to choose the optimal `K' of  K-means with the methods mentioned in Section \ref{sec-two}, since we cannot enumerate all the topics of length $K$, we can sample a set of  topics of length $K$, and use the sample to estimate the statistics of the population, of all the topics of length $K$.

\subsubsection{Using Partitioning Around Medoids (PAM)}
\begin{figure}
	\centering
	\includegraphics[width= \columnwidth]{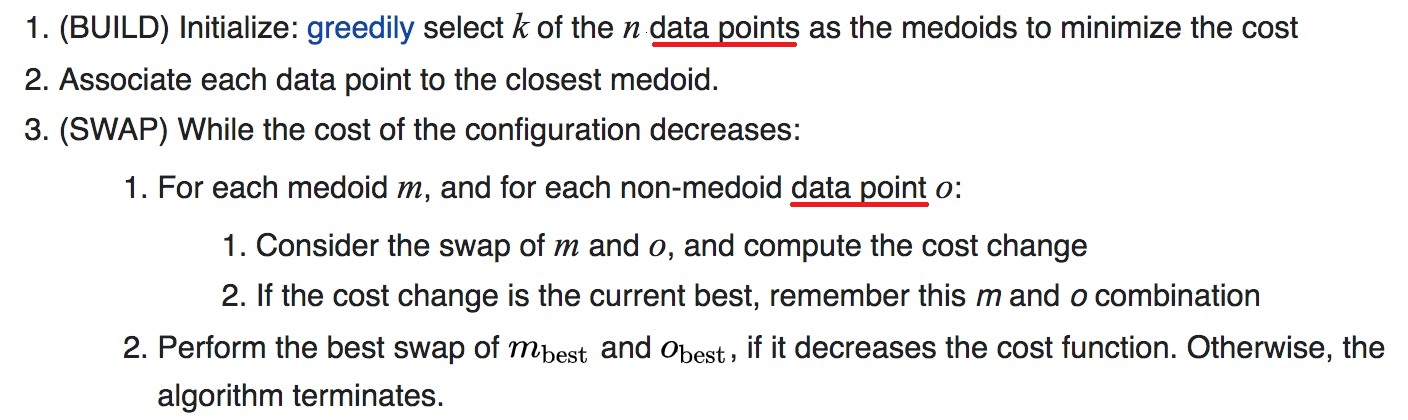}
	\caption{The PAM algorithm}
	\label{fig:pam}
\end{figure}

The PAM algorithm is widely used for finding  K-medoids \cite{kaufman1990partitioning}.  It can also be used for searching SCOM of length K (K-SCOMs) approximately, with some minor revisions. Figure \ref{fig:pam} shows how it works. The first revision is that a SCOM point does not need to be a data point; every node on the graph is a valid candidate. So we change the two positions on Figure \ref{fig:pam} where requiring ``data point'' to be ``any node on the graph''. The second revision is when calculating the cost, we need  considering the mass/weight of each data point, multiplying the squared distance by the mass.

Other clustering methods like Hierarchical clustering and Spectral clustering can also be used for finding K-SCOMs. We first  cluster the concepts in a document into $K$ groups, then search the SCOM-of-length-one of each group.

\subsection{Parallel to probabilistic topic models}
In Section \ref{sec-related-work}, we compared UM-S-TM with probabilistic topic models. Following is a parallel comparison between UM-S-TM and Probabilistic Topic Models (PTM).

\begin{itemize}
	\item In UM-S-TM, a concept that belongs to a topic is a sub-topic of the topic.  A sub-topic in UM-S-TM corresponds to a topic in PTM, which is a distribution over concepts.
	
	\item PTM has a hyper-parameter K, the number of topics, it can be optimized with techniques like Hierarchical Dirichlet Process (HDP) \cite{teh2006hierarchical}. The counterpart in UM-S-TM is the optimal $K$ of K-SCOMs, which can be calculated with the methods listed in Section \ref{sec-two}. Other methods for finding the optimal number of clusters, such as the Elbow method, the Silhouette method, or the Gap Statistic \cite{tibshirani2001estimating} may also be used.
	
	\item If the hyper-parameter K is fixed, in PTM, the optimal topics can be discovered with the Expectation-maximization algorithm, Gibbs Sampling \cite{porteous2008fast} , or variational inference; in UM-S-TM, the optimal K sub-topics of K-SCOMs can  be discovered with the methods listed in Section \ref{sec-approxi}. 
	
	\item In PTM, after the analysis, a document can be expressed as a distribution over the K topics. In UM-S-TM, a document can also be expressed as a distribution over the K sub-topics of K-SCOMs. Each concept in a document can be clustered according to its distance on the U-map, to each sub-topic of K-SCOMs. By counting the number of concepts in each cluster, the distribution is obtained.
\end{itemize}

Taking Doc1 supervised by U-map Deep as an example. After analysis of  UM-S-TM,
the optimal hyper-parameter K is 2;   the optimal 2-SCOMs is Topic $ \{P,R\}$.  It has two sub-topics: $P$ and $R$. Doc1 contains 12 concepts: $C, D, E, F, H, K, L, M, O, Q, R, S$, with term frequencies:  $12,  9,  13,  1,  5,  3,  22,  3,  3,  9,  10,  23$. Among them, Concepts $C ,F ,H ,M ,O ,Q$ are near sub-topic $P$ than $R$ (check U-map Deep); Concepts $D, E, K, L, R, S$ are near sub-topic $R$ than $P$. Therefore, Doc1 is clustered into two clusters. Cluster one's center is sub-topic $P$, it contains 33 concepts (12 + 1 + 5 + 3 + 3 + 9 = 33); Cluster two's center is sub-topic $R$, it contains 80 concepts (9 + 13 + 3 + 22 + 10 + 23 = 80). Therefore, Doc1 can be represented as a distribution: $[P:0.29, ~ R: 0.71]$.

\subsection{For information retrieval}
Since we can compare different documents' distance to a topic with Equation \ref{eq:document-to-topic-distance}, this property can be used for ranking a set of documents for  information retrieval.

If the Global SCOM of document $\tau$ is obtained, document $\tau$ can be expressed as a distribution over the K sub-topics of the Global SCOM. By comparing this distribution with the user's submitted keywords, we have another choice for ranking a set of documents. Following is one possible scoring method. 

In the method, the user is allowed to designate the proportion of her information needs. E.g., if the user submits keywords $[A:0.3,  ~  O: 0.7]$, that means the user wants to find a document that talks about Concept $A$ with proportion 0.3, and Concept $O$ with proportion 0.7. If the submitted keyword is $[E:1.0]$, that means the user wants to find a document that totally talks about Concept $E$.

\begin{definition}
Distance between distribution representation of information needs and distribution representation of Document $\tau$'s Global SCOM.
\begin{equation} 
d(\Theta, \Phi_\tau) = \sum_{i=1}^{M} \sum_{j=1}^{N} P_i Q_j d(C_i, C_j) 
\label{eq:information_retrieval}
\end{equation}
where $ \Theta $ is the distribution representation of information needs, there are $ M $ concepts in $ \Theta $, $ P_i $ is Concept $i$'s proportion in $ \Theta $;  $ \Phi_\tau $ is the distribution representation of Document $\tau$'s Global SCOM,  there are N concepts in $ \Phi_\tau $, $ Q_j $ is Concept $j$'s proportion in $ \Phi_\tau $; $ d(C_i, C_j) $ is Concept $i$ and $j$'s distance on the U-map.
\end{definition} 
Table \ref{tab:inf-retri}  lists Doc1's scores to different information needs, supervised by U-map Deep. Readers can check whether the scores are reasonable. The scores are the smaller the better.

\begin{table}
	\begin{center}
		\scalebox{0.99}{
			
\begin{tabular}{ll}
	\toprule
	Information needs & Scores \\  \midrule
	P:0.5, R:0.5      & 2.5    \\
	P:0.29, R:0.71    & 2.06   \\
	P:0.71, R:0.29    & 2.94   \\
	S:1.0             & 2.45   \\
	B:1.0             & 2.71   \\
	L:1.0             & 3.45   \\
	O:1.0             & 4.55   \\
	L:0.5, O:0.5      & 4      \\ \bottomrule
\end{tabular}

		}
\caption{Doc1's scores to different information needs, supervised by U-map Deep}
\label{tab:inf-retri}
	\end{center}
\end{table}

Another choice is to deem the information needs as a topic.
\begin{definition}
	Distance between topic representation of information needs and distribution representation of Document $\tau$'s Global SCOM.
	\begin{equation} 
	d(x, \Phi_\tau) =   \sum_{j=1}^{N}  Q_j d(x, C_j) 
	\label{eq:information_retrieval}
	\end{equation}
	where $ x $ is the topic representation of information needs;  $ \Phi_\tau $ is the distribution representation of Document $\tau$'s Global SCOM,  there are N concepts in $ \Phi_\tau $, $ Q_j $ is Concept $j$'s proportion in $ \Phi_\tau $; $ d(x, C_j) $ is concept-to-topic distance on the U-map.
\end{definition} 
In this scoring method, the user is not allowed to designate the proportion of her information needs.

\subsection{Using other semantic networks}
Other semantic networks such as Word Embedding  may also be used, as long as it can provide pair-wise distance information between concepts. Following is a method to calculate distance  between two concepts based on a collection of documents (a corpus) and Pointwise Mutual Information (PMI).

Pointwise Mutual Information is a measure of association used in information theory and statistics.\footnote{ \url{https://en.wikipedia.org/wiki/Pointwise_mutual_information}} 
\begin{equation*} 
{pmi} (x;y) = \log {\frac {p(x,y)}{p(x)p(y)}}=\log {\frac {p(x|y)}{p(x)}}=\log {\frac {p(y|x)}{p(y)}}
\label{eq:PMI}
\end{equation*}

\begin{definition}
Distance between two concepts based on PMI.
\begin{equation} 
d(x,y) = \frac {1}{e^{{pmi} (x;y)}}  = \frac {p(x)p(y)} {p(x,y)} =\frac{p(x)}  {p(x|y)}=\frac {p(y)} {p(y|x)}
\label{eq:PMI-dis}
\end{equation}
where $ x $ and $ y $ are two concepts, and $ x ~ != y $. $ p(x) $ and $ p(y) $ are probabilities of the concepts in the corpus. If there are $ N $ documents in the corpus, $ M $ of them contain Concept $ x $, then $ p(x) = M/N $. If $ T $ of them contain both $ x $ and $ y $, then   $ p(x,y) = T/N $. We assume $ p(x) $ and $ p(y) $ are positive.  The smallest distance between two concepts is zero, e.g., when  $ p(x) =  p(y) = p(x,y)  $ and all of them approaches zero. The largest distance between two concepts is infinity, e.g., when $ p(x) $ and  $ p(y) $ are positive,  and   $ p(x,y)  $ is zero. If $ x = y $, $ d(x,y) $ is defined to be 0.
\end{definition} 

With  Equation \ref{eq:PMI-dis}, we can construct a complete graph like Figure \ref{fig:complete-graph}. With the complete graph, distance between two concepts can be re-computed by finding the shortest path on the graph. Following is an example to show how it works.

Suppose we have a corpus which contains six documents: D1 to D6.  Table \ref{tab:corpus_distance} lists which concepts are contained in each document. Since term frequency is not used in the method, it is ignored in the table.

\begin{figure}
	\centering
	\includegraphics[width=0.9\columnwidth]{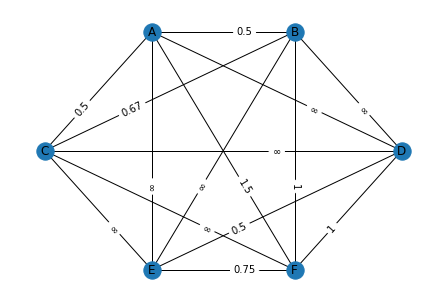}
	\caption{A complete graph calculated with a corpus}
	\label{fig:complete-dis}
\end{figure}

Table \ref{tab:probability-each} is the probability of each concept in the corpus. Table \ref{tab:Joint_probability} is the joint probability of each pair of concepts according to Table \ref{tab:corpus_distance}.  Table \ref{tab:Distances_concepts} is distances between each pair of concepts calculated with Equation \ref{eq:PMI-dis}. Figure \ref{fig:complete-dis} is the constructed complete graph from Table \ref{tab:Distances_concepts} ($ 2/3 $ is rounded to $ 0.67 $).
The complete graph can be simplified to look like an U-map, by removing redundant edges. An edge is defined to be redundant if there exists another path that has length less than or equal to the edge. E.g., in Figure \ref{fig:complete-dis}, the edge between Concept B and D has length infinity, there exists another path $B -->F -->D$, which has length 2 and is apparently less than infinity. Therefore, Edge $BD$ is redundant. Another example is the edge between Concept A and F, it has length $ 1.5 $, there exists another path $A -->B -->F$, which has length $ 1.5 $ and is equal to the length of Edge $ AF $. Therefore, Edge $AF$ is also redundant. The redundant edges should be located and removed dynamically, one after another.

\begin{table}
	\begin{center}
		\scalebox{0.99}{

\begin{tabular}{ll}
	\toprule
	Doc & Concepts contained \\  \midrule
	D1  & A, B, C            \\
	D2  & A, B, F            \\
	D3  & A, C               \\
	D4  & D, E, F            \\
	D5  & D, E               \\
	D6  & E, F              \\ \bottomrule
\end{tabular}		
					
		}
\caption{A corpus for constructing a semantic network}
\label{tab:corpus_distance}
	\end{center}
\end{table}
\begin{table}
	\begin{center}
		\scalebox{0.99}{

\begin{tabular}{llllll}
		\toprule
	A   & B   & C   & D   & E   & F  \\  \midrule
	1/2 & 1/3 & 1/3 & 1/3 & 1/2 & 1/2  \\ \bottomrule
\end{tabular}
			
		}
\caption{Probability of each concept in the corpus}
\label{tab:probability-each}
	\end{center}
\end{table}
\begin{table}
	\begin{center}
		\scalebox{0.99}{		
\begin{tabular}{lllllll}
	\toprule
	& A & B   & C   & D & E   & F   \\  \midrule
	A &   & 1/3 & 1/3 & 0 & 0   & 1/6 \\
	B &   &     & 1/6 & 0 & 0   & 1/6 \\
	C &   &     &     & 0 & 0   & 0   \\
	D &   &     &     &   & 1/3 & 1/6 \\
	E &   &     &     &   &     & 1/3 \\
	F &   &     &     &   &     &     \\ \bottomrule
\end{tabular}					
		}
\caption{Joint probability of each pair of concepts}
\label{tab:Joint_probability}
	\end{center}
\end{table}
\begin{table}
	\begin{center}
		\scalebox{0.99}{

			\begin{tabular}{lllllll}
					\toprule
				& A & B   & C   & D   & E   & F   \\ \midrule
				A & 0 & 1/2 & 1/2 & $\infty$ & $\infty$ & 3/2 \\
				B &   & 0   & 2/3 & $\infty$ & $\infty$ & 1   \\
				C &   &     & 0   & $\infty$ & $\infty$ & $\infty$ \\
				D &   &     &     & 0   & 1/2 & 1   \\
				E &   &     &     &     & 0   & 3/4 \\
				F &   &     &     &     &     & 0  \\ \bottomrule
			\end{tabular}

		}
	
\caption{Distances calculated with Equation \ref{eq:PMI-dis}}
\label{tab:Distances_concepts}
	\end{center}
\end{table}

\begin{figure}
	\centering
	\includegraphics[width=0.9\columnwidth]{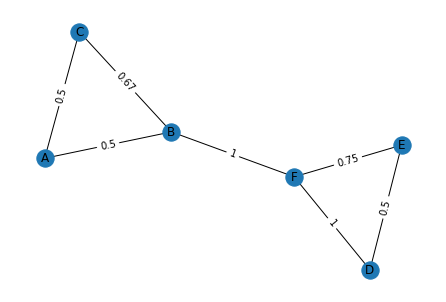}
	\caption{A constructed semantic network from a corpus}
	\label{fig:simplified-graph}
\end{figure}

Figure \ref{fig:simplified-graph} is the resulting semantic network by simplifying the complete graph. With the  network, we can re-compute distances between concepts as the shortest path on the graph.

\subsection{Curve fitting  local champions only}
A further examination shows that, for the curve fitting method discussed  in Section \ref{sec-curve-fitting}, the population information of each group seems not necessary; we can fit a curve on the scores of local champions, and let the curve deal with the differences between groups. That is, in Equation \ref{equ:three-step1}, let $ z(x_n^\star) =  d(\tau, x_n^\star) $, and ignore the  mean and standard deviation of each group, the normalization factor $\phi_x$ for each group can also be ignored.  

\begin{figure}
	\centering
	\includegraphics[width=0.9\columnwidth]{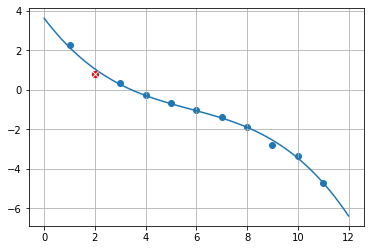}
	\caption{Curve fitting $\log  d(\tau, x_n^\star) $, Doc1 supervised by U-map Deep}
	\label{fig:curve_of_logW}
\end{figure}

We can also use other scoring method, such as $ z(x_n^\star) = \log  d(\tau, x_n^\star) $. Figure \ref{fig:curve_of_logW} is the result of curve fitting $\log  d(\tau, x_n^\star) $ when Doc1 is supervised by U-map Deep. The curve is selected by Cross Validation and Grid Search. Normalization factor $\phi_x$ is not used for calculating $\log  d(\tau, x_n^\star) $. The curve also suggests that Topic $\{P,R\}$  is the SCOM of Doc1 supervised by U-map Deep. Since the logarithm function cannot deal with 0, topics with length greater than or equal to the document's concept-length are excluded from the candidate topic set.

A further exploration is to use the curve fitting method to select the optimal number of topics in probabilistic topic models, or the optimal number of clusters in clustering. The score for each local champion could be the minus-average-log-likelihood of all the concepts in a corpus. The ``minus'' is to make sure that the criterion is ``the smaller the better''. For selecting the optimal number of clusters problem, the criterion could be logarithm of average-within-cluster-variance.

\subsection{For graph partition}
\begin{figure}
	\centering
	\includegraphics[width=0.9\columnwidth]{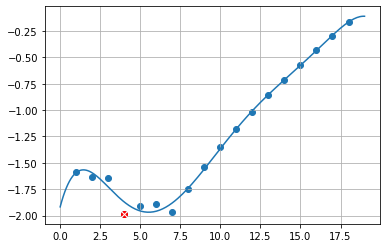}
	\caption{Graph partition of U-map Shallow by UM-S-TM}
	\label{fig:Graph_partition_shallow}
\end{figure}

In mathematics, a graph partition is the reduction of a graph to a smaller graph by partitioning its set of nodes into mutually exclusive groups \cite{Bourse2014, Wang2014}.
For each U-map, there is a special document that contains all the concepts in the U-map
and all the term frequencies equals exactly one. By analyzing this special document with 
UM-S-TM, a partition of the graph is obtained. E.g., the curve fitted in Figure \ref{fig:Graph_partition_shallow}  suggests U-map Shallow should be partitioned into four clusters. The score for each local champion is calculated with Equation \ref{equ:three-step1}. However, for this special document, the standard deviation (SD) of the last group of topics (that is when topic length equals 19) is zero. To avoid division by zero, the last group of topics are excluded from the candidate topic set.

The four centers of the clusters are $\{A, F , G, N \}$ or $\{E, F , G, N \}$, they get the same score. By comparing each node's distance to each center, partition of the graph is obtained.

\subsection{Analyzing images}
If we consider an image as a ``bag-of-colors'', then we can use UM-S-TM to analyze the SCOM of an image. Then use the SCOMs to search or cluster images. Note the pairwise distances between colors are easy to compute, because we have the coordinates for each color.

\subsection{Differentiating directions }
By far we are ignoring the directions of an U-map. Maybe more information can be extracted by differentiating directions on U-map. E.g., if there exists a link from Concept A to Concept B, defining the distance from A to B to be one, and from B to A to be $1+\epsilon$, where $\epsilon$ can be a small positive or negative number. If so, the first six rows (seven rows if the head row is counted) of Table \ref{tab:appA1} is not a symmetric matrix any more.

\subsection{Multiple Tutors Supervised Topic Model}
In Multiple Tutors Supervised Topic Model (MT-S-TM), the final SCOM is determined by multiple supervisors. The first supervisor is an U-map, it decides the SCOM can only be selected from the set of local champions, like the ones listed in the ``Topic'' column of Table \ref{tab:3-steps-table}. The second supervisor is a training data set. E.g., a collection of documents with topics tagged by a human or a probabilistic topic model. We use the training data set to train an optimal parameter of $\alpha$ or $\delta$ in Equation \ref{mass-dis-fix32} and \ref{mass-dis-fix4}, or a hyper-parameter for curve-fitting in Method 1, then use the trained parameter to select the final SCOM, from the set of local champions.

The training process can be implemented with gradient decent. Taking training parameter  $\alpha$ as an example, we first initialize $\alpha$ with some random value, then calculate SCOMs for the training documents with Equation \ref{mass-dis-fix32}, then calculate the sum of distances between the returned SCOMs and the tagged topics, it is the objective to be minimized, then adjust $\alpha$ to decrease the objective.

\section{Conclusion}
In this paper, we propose a model called Understanding Map Supervised Topic Model (UM-S-TM). The aim of this model is to discover the ``topic'' of a document. The topic is decided both by the content of the document, and supervised by a semantic network, specifically, Understanding Map. Inspired by the notion of Center of Mass in physics, an extension called Semantic Center of Mass (SCOM) is proposed, and deemed as the abstract ``topic'' of a document. Based on different justifications, three possible methods are devised to discover the SCOM. Some experiments on artificial documents and U-maps are conducted to test their outcomes. Evidence shows there seems exist a special topic associated with a document supervised by an U-map, such that 97\% of test cases, at least two of the three methods have an agreement on the discovered SCOM. 68\% of test cases, the three methods have unanimous agreement on the discovered SCOM. We also compared UM-S-TM vectorization of documents with term frequency vectorization, and its ability of capturing sequential information.

\bibliographystyle{ACM-Reference-Format}
\bibliography{uuu} 

\clearpage
\appendix

\begin{table}
	\begin{center}
		\scalebox{0.8}{
			
			\begin{tabular}{llrrrrrr}
				\toprule
				{} &  Topic &    A &     B &    C &    D &    E &     F \\
				\midrule
				0  &      A &  0.0 &   1.0 &  1.0 &  4.0 &  4.0 &   9.0 \\
				1  &      B &  1.0 &   0.0 &  4.0 &  9.0 &  9.0 &  16.0 \\
				2  &      C &  1.0 &   4.0 &  0.0 &  1.0 &  1.0 &   4.0 \\
				3  &      D &  4.0 &   9.0 &  1.0 &  0.0 &  1.0 &   1.0 \\
				4  &      E &  4.0 &   9.0 &  1.0 &  1.0 &  0.0 &   4.0 \\
				5  &      F &  9.0 &  16.0 &  4.0 &  1.0 &  4.0 &   0.0 \\
				6  &     AB &  0.0 &   0.0 &  1.0 &  4.0 &  4.0 &   9.0 \\
				7  &     AC &  0.0 &   1.0 &  0.0 &  1.0 &  1.0 &   4.0 \\
				8  &     AD &  0.0 &   1.0 &  1.0 &  0.0 &  1.0 &   1.0 \\
				9  &     AE &  0.0 &   1.0 &  1.0 &  1.0 &  0.0 &   4.0 \\
				10 &     AF &  0.0 &   1.0 &  1.0 &  1.0 &  4.0 &   0.0 \\
				11 &     BC &  1.0 &   0.0 &  0.0 &  1.0 &  1.0 &   4.0 \\
				12 &     BD &  1.0 &   0.0 &  1.0 &  0.0 &  1.0 &   1.0 \\
				13 &     BE &  1.0 &   0.0 &  1.0 &  1.0 &  0.0 &   4.0 \\
				14 &     BF &  1.0 &   0.0 &  4.0 &  1.0 &  4.0 &   0.0 \\
				15 &     CD &  1.0 &   4.0 &  0.0 &  0.0 &  1.0 &   1.0 \\
				16 &     CE &  1.0 &   4.0 &  0.0 &  1.0 &  0.0 &   4.0 \\
				17 &     CF &  1.0 &   4.0 &  0.0 &  1.0 &  1.0 &   0.0 \\
				18 &     DE &  4.0 &   9.0 &  1.0 &  0.0 &  0.0 &   1.0 \\
				19 &     DF &  4.0 &   9.0 &  1.0 &  0.0 &  1.0 &   0.0 \\
				20 &     EF &  4.0 &   9.0 &  1.0 &  1.0 &  0.0 &   0.0 \\
				21 &    ABC &  0.0 &   0.0 &  0.0 &  1.0 &  1.0 &   4.0 \\
				22 &    ABD &  0.0 &   0.0 &  1.0 &  0.0 &  1.0 &   1.0 \\
				23 &    ABE &  0.0 &   0.0 &  1.0 &  1.0 &  0.0 &   4.0 \\
				24 &    ABF &  0.0 &   0.0 &  1.0 &  1.0 &  4.0 &   0.0 \\
				25 &    ACD &  0.0 &   1.0 &  0.0 &  0.0 &  1.0 &   1.0 \\
				26 &    ACE &  0.0 &   1.0 &  0.0 &  1.0 &  0.0 &   4.0 \\
				27 &    ACF &  0.0 &   1.0 &  0.0 &  1.0 &  1.0 &   0.0 \\
				28 &    ADE &  0.0 &   1.0 &  1.0 &  0.0 &  0.0 &   1.0 \\
				29 &    ADF &  0.0 &   1.0 &  1.0 &  0.0 &  1.0 &   0.0 \\
				30 &    AEF &  0.0 &   1.0 &  1.0 &  1.0 &  0.0 &   0.0 \\
				31 &    BCD &  1.0 &   0.0 &  0.0 &  0.0 &  1.0 &   1.0 \\
				32 &    BCE &  1.0 &   0.0 &  0.0 &  1.0 &  0.0 &   4.0 \\
				33 &    BCF &  1.0 &   0.0 &  0.0 &  1.0 &  1.0 &   0.0 \\
				34 &    BDE &  1.0 &   0.0 &  1.0 &  0.0 &  0.0 &   1.0 \\
				35 &    BDF &  1.0 &   0.0 &  1.0 &  0.0 &  1.0 &   0.0 \\
				36 &    BEF &  1.0 &   0.0 &  1.0 &  1.0 &  0.0 &   0.0 \\
				37 &    CDE &  1.0 &   4.0 &  0.0 &  0.0 &  0.0 &   1.0 \\
				38 &    CDF &  1.0 &   4.0 &  0.0 &  0.0 &  1.0 &   0.0 \\
				39 &    CEF &  1.0 &   4.0 &  0.0 &  1.0 &  0.0 &   0.0 \\
				40 &    DEF &  4.0 &   9.0 &  1.0 &  0.0 &  0.0 &   0.0 \\
				41 &   ABCD &  0.0 &   0.0 &  0.0 &  0.0 &  1.0 &   1.0 \\
				42 &   ABCE &  0.0 &   0.0 &  0.0 &  1.0 &  0.0 &   4.0 \\
				43 &   ABCF &  0.0 &   0.0 &  0.0 &  1.0 &  1.0 &   0.0 \\
				44 &   ABDE &  0.0 &   0.0 &  1.0 &  0.0 &  0.0 &   1.0 \\
				45 &   ABDF &  0.0 &   0.0 &  1.0 &  0.0 &  1.0 &   0.0 \\
				46 &   ABEF &  0.0 &   0.0 &  1.0 &  1.0 &  0.0 &   0.0 \\
				47 &   ACDE &  0.0 &   1.0 &  0.0 &  0.0 &  0.0 &   1.0 \\
				48 &   ACDF &  0.0 &   1.0 &  0.0 &  0.0 &  1.0 &   0.0 \\
				49 &   ACEF &  0.0 &   1.0 &  0.0 &  1.0 &  0.0 &   0.0 \\
				50 &   ADEF &  0.0 &   1.0 &  1.0 &  0.0 &  0.0 &   0.0 \\
				51 &   BCDE &  1.0 &   0.0 &  0.0 &  0.0 &  0.0 &   1.0 \\
				52 &   BCDF &  1.0 &   0.0 &  0.0 &  0.0 &  1.0 &   0.0 \\
				53 &   BCEF &  1.0 &   0.0 &  0.0 &  1.0 &  0.0 &   0.0 \\
				54 &   BDEF &  1.0 &   0.0 &  1.0 &  0.0 &  0.0 &   0.0 \\
				55 &   CDEF &  1.0 &   4.0 &  0.0 &  0.0 &  0.0 &   0.0 \\
				56 &  ABCDE &  0.0 &   0.0 &  0.0 &  0.0 &  0.0 &   1.0 \\
				57 &  ABCDF &  0.0 &   0.0 &  0.0 &  0.0 &  1.0 &   0.0 \\
				58 &  ABCEF &  0.0 &   0.0 &  0.0 &  1.0 &  0.0 &   0.0 \\
				59 &  ABDEF &  0.0 &   0.0 &  1.0 &  0.0 &  0.0 &   0.0 \\
				60 &  ACDEF &  0.0 &   1.0 &  0.0 &  0.0 &  0.0 &   0.0 \\
				61 &  BCDEF &  1.0 &   0.0 &  0.0 &  0.0 &  0.0 &   0.0 \\
				\bottomrule
			\end{tabular}}
\caption{Squared concept-to-topic distances of U-map One}
\label{tab:appA1}
\end{center}
\end{table}

\begin{table}
	\begin{center}
		\scalebox{0.8}{
			
		\begin{tabular}{llrrrrrr}
			\toprule
			{} &  Topic &      A &      B &      C &      D &      E &      F \\
			\midrule
			0  &      A &   0.00 &   1.00 &   1.00 &   4.00 &   4.00 &   9.00 \\
			1  &      B &   1.00 &   0.00 &   4.00 &   9.00 &   9.00 &  16.00 \\
			2  &      C &   1.00 &   4.00 &   0.00 &   1.00 &   1.00 &   4.00 \\
			3  &      D &   4.00 &   9.00 &   1.00 &   0.00 &   1.00 &   1.00 \\
			4  &      E &   4.00 &   9.00 &   1.00 &   1.00 &   0.00 &   4.00 \\
			5  &      F &   9.00 &  16.00 &   4.00 &   1.00 &   4.00 &   0.00 \\
			6  &     AB &   0.00 &   0.00 &   2.46 &   9.86 &   9.86 &  22.18 \\
			7  &     AC &   0.00 &   2.46 &   0.00 &   2.46 &   2.46 &   9.86 \\
			8  &     AD &   0.00 &   2.46 &   2.46 &   0.00 &   2.46 &   2.46 \\
			9  &     AE &   0.00 &   2.46 &   2.46 &   2.46 &   0.00 &   9.86 \\
			10 &     AF &   0.00 &   2.46 &   2.46 &   2.46 &   9.86 &   0.00 \\
			11 &     BC &   2.46 &   0.00 &   0.00 &   2.46 &   2.46 &   9.86 \\
			12 &     BD &   2.46 &   0.00 &   2.46 &   0.00 &   2.46 &   2.46 \\
			13 &     BE &   2.46 &   0.00 &   2.46 &   2.46 &   0.00 &   9.86 \\
			14 &     BF &   2.46 &   0.00 &   9.86 &   2.46 &   9.86 &   0.00 \\
			15 &     CD &   2.46 &   9.86 &   0.00 &   0.00 &   2.46 &   2.46 \\
			16 &     CE &   2.46 &   9.86 &   0.00 &   2.46 &   0.00 &   9.86 \\
			17 &     CF &   2.46 &   9.86 &   0.00 &   2.46 &   2.46 &   0.00 \\
			18 &     DE &   9.86 &  22.18 &   2.46 &   0.00 &   0.00 &   2.46 \\
			19 &     DF &   9.86 &  22.18 &   2.46 &   0.00 &   2.46 &   0.00 \\
			20 &     EF &   9.86 &  22.18 &   2.46 &   2.46 &   0.00 &   0.00 \\
			21 &    ABC &   0.00 &   0.00 &   0.00 &   4.84 &   4.84 &  19.37 \\
			22 &    ABD &   0.00 &   0.00 &   4.84 &   0.00 &   4.84 &   4.84 \\
			23 &    ABE &   0.00 &   0.00 &   4.84 &   4.84 &   0.00 &  19.37 \\
			24 &    ABF &   0.00 &   0.00 &   4.84 &   4.84 &  19.37 &   0.00 \\
			25 &    ACD &   0.00 &   4.84 &   0.00 &   0.00 &   4.84 &   4.84 \\
			26 &    ACE &   0.00 &   4.84 &   0.00 &   4.84 &   0.00 &  19.37 \\
			27 &    ACF &   0.00 &   4.84 &   0.00 &   4.84 &   4.84 &   0.00 \\
			28 &    ADE &   0.00 &   4.84 &   4.84 &   0.00 &   0.00 &   4.84 \\
			29 &    ADF &   0.00 &   4.84 &   4.84 &   0.00 &   4.84 &   0.00 \\
			30 &    AEF &   0.00 &   4.84 &   4.84 &   4.84 &   0.00 &   0.00 \\
			31 &    BCD &   4.84 &   0.00 &   0.00 &   0.00 &   4.84 &   4.84 \\
			32 &    BCE &   4.84 &   0.00 &   0.00 &   4.84 &   0.00 &  19.37 \\
			33 &    BCF &   4.84 &   0.00 &   0.00 &   4.84 &   4.84 &   0.00 \\
			34 &    BDE &   4.84 &   0.00 &   4.84 &   0.00 &   0.00 &   4.84 \\
			35 &    BDF &   4.84 &   0.00 &   4.84 &   0.00 &   4.84 &   0.00 \\
			36 &    BEF &   4.84 &   0.00 &   4.84 &   4.84 &   0.00 &   0.00 \\
			37 &    CDE &   4.84 &  19.37 &   0.00 &   0.00 &   0.00 &   4.84 \\
			38 &    CDF &   4.84 &  19.37 &   0.00 &   0.00 &   4.84 &   0.00 \\
			39 &    CEF &   4.84 &  19.37 &   0.00 &   4.84 &   0.00 &   0.00 \\
			40 &    DEF &  19.37 &  43.58 &   4.84 &   0.00 &   0.00 &   0.00 \\
			41 &   ABCD &   0.00 &   0.00 &   0.00 &   0.00 &   9.58 &   9.58 \\
			42 &   ABCE &   0.00 &   0.00 &   0.00 &   9.58 &   0.00 &  38.33 \\
			43 &   ABCF &   0.00 &   0.00 &   0.00 &   9.58 &   9.58 &   0.00 \\
			44 &   ABDE &   0.00 &   0.00 &   9.58 &   0.00 &   0.00 &   9.58 \\
			45 &   ABDF &   0.00 &   0.00 &   9.58 &   0.00 &   9.58 &   0.00 \\
			46 &   ABEF &   0.00 &   0.00 &   9.58 &   9.58 &   0.00 &   0.00 \\
			47 &   ACDE &   0.00 &   9.58 &   0.00 &   0.00 &   0.00 &   9.58 \\
			48 &   ACDF &   0.00 &   9.58 &   0.00 &   0.00 &   9.58 &   0.00 \\
			49 &   ACEF &   0.00 &   9.58 &   0.00 &   9.58 &   0.00 &   0.00 \\
			50 &   ADEF &   0.00 &   9.58 &   9.58 &   0.00 &   0.00 &   0.00 \\
			51 &   BCDE &   9.58 &   0.00 &   0.00 &   0.00 &   0.00 &   9.58 \\
			52 &   BCDF &   9.58 &   0.00 &   0.00 &   0.00 &   9.58 &   0.00 \\
			53 &   BCEF &   9.58 &   0.00 &   0.00 &   9.58 &   0.00 &   0.00 \\
			54 &   BDEF &   9.58 &   0.00 &   9.58 &   0.00 &   0.00 &   0.00 \\
			55 &   CDEF &   9.58 &  38.33 &   0.00 &   0.00 &   0.00 &   0.00 \\
			56 &  ABCDE &   0.00 &   0.00 &   0.00 &   0.00 &   0.00 &  23.00 \\
			57 &  ABCDF &   0.00 &   0.00 &   0.00 &   0.00 &  23.00 &   0.00 \\
			58 &  ABCEF &   0.00 &   0.00 &   0.00 &  23.00 &   0.00 &   0.00 \\
			59 &  ABDEF &   0.00 &   0.00 &  23.00 &   0.00 &   0.00 &   0.00 \\
			60 &  ACDEF &   0.00 &  23.00 &   0.00 &   0.00 &   0.00 &   0.00 \\
			61 &  BCDEF &  23.00 &   0.00 &   0.00 &   0.00 &   0.00 &   0.00 \\
			\bottomrule
		\end{tabular} }
\caption{Normalized squared distances of U-map One}
\label{tab:appA2}
	\end{center}
\end{table}

\begin{table}
	\begin{center}
		\scalebox{0.8}{
			
			\begin{tabular}{llrrrrrr}
				\toprule
				{} &  Topic &    A &     B &    C &    D &    E &     F \\
				\midrule
				0  &      A &  0.2 &   1.2 &  1.2 &  4.2 &  4.2 &   9.2 \\
				1  &      B &  1.2 &   0.2 &  4.2 &  9.2 &  9.2 &  16.2 \\
				2  &      C &  1.2 &   4.2 &  0.2 &  1.2 &  1.2 &   4.2 \\
				3  &      D &  4.2 &   9.2 &  1.2 &  0.2 &  1.2 &   1.2 \\
				4  &      E &  4.2 &   9.2 &  1.2 &  1.2 &  0.2 &   4.2 \\
				5  &      F &  9.2 &  16.2 &  4.2 &  1.2 &  4.2 &   0.2 \\
				6  &     AB &  0.2 &   0.2 &  1.2 &  4.2 &  4.2 &   9.2 \\
				7  &     AC &  0.2 &   1.2 &  0.2 &  1.2 &  1.2 &   4.2 \\
				8  &     AD &  0.2 &   1.2 &  1.2 &  0.2 &  1.2 &   1.2 \\
				9  &     AE &  0.2 &   1.2 &  1.2 &  1.2 &  0.2 &   4.2 \\
				10 &     AF &  0.2 &   1.2 &  1.2 &  1.2 &  4.2 &   0.2 \\
				11 &     BC &  1.2 &   0.2 &  0.2 &  1.2 &  1.2 &   4.2 \\
				12 &     BD &  1.2 &   0.2 &  1.2 &  0.2 &  1.2 &   1.2 \\
				13 &     BE &  1.2 &   0.2 &  1.2 &  1.2 &  0.2 &   4.2 \\
				14 &     BF &  1.2 &   0.2 &  4.2 &  1.2 &  4.2 &   0.2 \\
				15 &     CD &  1.2 &   4.2 &  0.2 &  0.2 &  1.2 &   1.2 \\
				16 &     CE &  1.2 &   4.2 &  0.2 &  1.2 &  0.2 &   4.2 \\
				17 &     CF &  1.2 &   4.2 &  0.2 &  1.2 &  1.2 &   0.2 \\
				18 &     DE &  4.2 &   9.2 &  1.2 &  0.2 &  0.2 &   1.2 \\
				19 &     DF &  4.2 &   9.2 &  1.2 &  0.2 &  1.2 &   0.2 \\
				20 &     EF &  4.2 &   9.2 &  1.2 &  1.2 &  0.2 &   0.2 \\
				21 &    ABC &  0.2 &   0.2 &  0.2 &  1.2 &  1.2 &   4.2 \\
				22 &    ABD &  0.2 &   0.2 &  1.2 &  0.2 &  1.2 &   1.2 \\
				23 &    ABE &  0.2 &   0.2 &  1.2 &  1.2 &  0.2 &   4.2 \\
				24 &    ABF &  0.2 &   0.2 &  1.2 &  1.2 &  4.2 &   0.2 \\
				25 &    ACD &  0.2 &   1.2 &  0.2 &  0.2 &  1.2 &   1.2 \\
				26 &    ACE &  0.2 &   1.2 &  0.2 &  1.2 &  0.2 &   4.2 \\
				27 &    ACF &  0.2 &   1.2 &  0.2 &  1.2 &  1.2 &   0.2 \\
				28 &    ADE &  0.2 &   1.2 &  1.2 &  0.2 &  0.2 &   1.2 \\
				29 &    ADF &  0.2 &   1.2 &  1.2 &  0.2 &  1.2 &   0.2 \\
				30 &    AEF &  0.2 &   1.2 &  1.2 &  1.2 &  0.2 &   0.2 \\
				31 &    BCD &  1.2 &   0.2 &  0.2 &  0.2 &  1.2 &   1.2 \\
				32 &    BCE &  1.2 &   0.2 &  0.2 &  1.2 &  0.2 &   4.2 \\
				33 &    BCF &  1.2 &   0.2 &  0.2 &  1.2 &  1.2 &   0.2 \\
				34 &    BDE &  1.2 &   0.2 &  1.2 &  0.2 &  0.2 &   1.2 \\
				35 &    BDF &  1.2 &   0.2 &  1.2 &  0.2 &  1.2 &   0.2 \\
				36 &    BEF &  1.2 &   0.2 &  1.2 &  1.2 &  0.2 &   0.2 \\
				37 &    CDE &  1.2 &   4.2 &  0.2 &  0.2 &  0.2 &   1.2 \\
				38 &    CDF &  1.2 &   4.2 &  0.2 &  0.2 &  1.2 &   0.2 \\
				39 &    CEF &  1.2 &   4.2 &  0.2 &  1.2 &  0.2 &   0.2 \\
				40 &    DEF &  4.2 &   9.2 &  1.2 &  0.2 &  0.2 &   0.2 \\
				41 &   ABCD &  0.2 &   0.2 &  0.2 &  0.2 &  1.2 &   1.2 \\
				42 &   ABCE &  0.2 &   0.2 &  0.2 &  1.2 &  0.2 &   4.2 \\
				43 &   ABCF &  0.2 &   0.2 &  0.2 &  1.2 &  1.2 &   0.2 \\
				44 &   ABDE &  0.2 &   0.2 &  1.2 &  0.2 &  0.2 &   1.2 \\
				45 &   ABDF &  0.2 &   0.2 &  1.2 &  0.2 &  1.2 &   0.2 \\
				46 &   ABEF &  0.2 &   0.2 &  1.2 &  1.2 &  0.2 &   0.2 \\
				47 &   ACDE &  0.2 &   1.2 &  0.2 &  0.2 &  0.2 &   1.2 \\
				48 &   ACDF &  0.2 &   1.2 &  0.2 &  0.2 &  1.2 &   0.2 \\
				49 &   ACEF &  0.2 &   1.2 &  0.2 &  1.2 &  0.2 &   0.2 \\
				50 &   ADEF &  0.2 &   1.2 &  1.2 &  0.2 &  0.2 &   0.2 \\
				51 &   BCDE &  1.2 &   0.2 &  0.2 &  0.2 &  0.2 &   1.2 \\
				52 &   BCDF &  1.2 &   0.2 &  0.2 &  0.2 &  1.2 &   0.2 \\
				53 &   BCEF &  1.2 &   0.2 &  0.2 &  1.2 &  0.2 &   0.2 \\
				54 &   BDEF &  1.2 &   0.2 &  1.2 &  0.2 &  0.2 &   0.2 \\
				55 &   CDEF &  1.2 &   4.2 &  0.2 &  0.2 &  0.2 &   0.2 \\
				56 &  ABCDE &  0.2 &   0.2 &  0.2 &  0.2 &  0.2 &   1.2 \\
				57 &  ABCDF &  0.2 &   0.2 &  0.2 &  0.2 &  1.2 &   0.2 \\
				58 &  ABCEF &  0.2 &   0.2 &  0.2 &  1.2 &  0.2 &   0.2 \\
				59 &  ABDEF &  0.2 &   0.2 &  1.2 &  0.2 &  0.2 &   0.2 \\
				60 &  ACDEF &  0.2 &   1.2 &  0.2 &  0.2 &  0.2 &   0.2 \\
				61 &  BCDEF &  1.2 &   0.2 &  0.2 &  0.2 &  0.2 &   0.2 \\
				\bottomrule
			\end{tabular}
			
		}
\caption{Squared distances of U-map One with 0.2 noise}
\label{tab:appA3}
	\end{center}
\end{table}

\begin{table}
	\begin{center}
		\scalebox{0.8}{
			
	\begin{tabular}{llrrrrrr}
		\toprule
		{} &  Topic &      A &      B &      C &      D &      E &      F \\
		\midrule
		0  &      A &   0.20 &   1.20 &   1.20 &   4.20 &   4.20 &   9.20 \\
		1  &      B &   1.20 &   0.20 &   4.20 &   9.20 &   9.20 &  16.20 \\
		2  &      C &   1.20 &   4.20 &   0.20 &   1.20 &   1.20 &   4.20 \\
		3  &      D &   4.20 &   9.20 &   1.20 &   0.20 &   1.20 &   1.20 \\
		4  &      E &   4.20 &   9.20 &   1.20 &   1.20 &   0.20 &   4.20 \\
		5  &      F &   9.20 &  16.20 &   4.20 &   1.20 &   4.20 &   0.20 \\
		6  &     AB &   0.46 &   0.46 &   2.76 &   9.65 &   9.65 &  21.14 \\
		7  &     AC &   0.46 &   2.76 &   0.46 &   2.76 &   2.76 &   9.65 \\
		8  &     AD &   0.46 &   2.76 &   2.76 &   0.46 &   2.76 &   2.76 \\
		9  &     AE &   0.46 &   2.76 &   2.76 &   2.76 &   0.46 &   9.65 \\
		10 &     AF &   0.46 &   2.76 &   2.76 &   2.76 &   9.65 &   0.46 \\
		11 &     BC &   2.76 &   0.46 &   0.46 &   2.76 &   2.76 &   9.65 \\
		12 &     BD &   2.76 &   0.46 &   2.76 &   0.46 &   2.76 &   2.76 \\
		13 &     BE &   2.76 &   0.46 &   2.76 &   2.76 &   0.46 &   9.65 \\
		14 &     BF &   2.76 &   0.46 &   9.65 &   2.76 &   9.65 &   0.46 \\
		15 &     CD &   2.76 &   9.65 &   0.46 &   0.46 &   2.76 &   2.76 \\
		16 &     CE &   2.76 &   9.65 &   0.46 &   2.76 &   0.46 &   9.65 \\
		17 &     CF &   2.76 &   9.65 &   0.46 &   2.76 &   2.76 &   0.46 \\
		18 &     DE &   9.65 &  21.14 &   2.76 &   0.46 &   0.46 &   2.76 \\
		19 &     DF &   9.65 &  21.14 &   2.76 &   0.46 &   2.76 &   0.46 \\
		20 &     EF &   9.65 &  21.14 &   2.76 &   2.76 &   0.46 &   0.46 \\
		21 &    ABC &   0.81 &   0.81 &   0.81 &   4.88 &   4.88 &  17.08 \\
		22 &    ABD &   0.81 &   0.81 &   4.88 &   0.81 &   4.88 &   4.88 \\
		23 &    ABE &   0.81 &   0.81 &   4.88 &   4.88 &   0.81 &  17.08 \\
		24 &    ABF &   0.81 &   0.81 &   4.88 &   4.88 &  17.08 &   0.81 \\
		25 &    ACD &   0.81 &   4.88 &   0.81 &   0.81 &   4.88 &   4.88 \\
		26 &    ACE &   0.81 &   4.88 &   0.81 &   4.88 &   0.81 &  17.08 \\
		27 &    ACF &   0.81 &   4.88 &   0.81 &   4.88 &   4.88 &   0.81 \\
		28 &    ADE &   0.81 &   4.88 &   4.88 &   0.81 &   0.81 &   4.88 \\
		29 &    ADF &   0.81 &   4.88 &   4.88 &   0.81 &   4.88 &   0.81 \\
		30 &    AEF &   0.81 &   4.88 &   4.88 &   4.88 &   0.81 &   0.81 \\
		31 &    BCD &   4.88 &   0.81 &   0.81 &   0.81 &   4.88 &   4.88 \\
		32 &    BCE &   4.88 &   0.81 &   0.81 &   4.88 &   0.81 &  17.08 \\
		33 &    BCF &   4.88 &   0.81 &   0.81 &   4.88 &   4.88 &   0.81 \\
		34 &    BDE &   4.88 &   0.81 &   4.88 &   0.81 &   0.81 &   4.88 \\
		35 &    BDF &   4.88 &   0.81 &   4.88 &   0.81 &   4.88 &   0.81 \\
		36 &    BEF &   4.88 &   0.81 &   4.88 &   4.88 &   0.81 &   0.81 \\
		37 &    CDE &   4.88 &  17.08 &   0.81 &   0.81 &   0.81 &   4.88 \\
		38 &    CDF &   4.88 &  17.08 &   0.81 &   0.81 &   4.88 &   0.81 \\
		39 &    CEF &   4.88 &  17.08 &   0.81 &   4.88 &   0.81 &   0.81 \\
		40 &    DEF &  17.08 &  37.42 &   4.88 &   0.81 &   0.81 &   0.81 \\
		41 &   ABCD &   1.34 &   1.34 &   1.34 &   1.34 &   8.07 &   8.07 \\
		42 &   ABCE &   1.34 &   1.34 &   1.34 &   8.07 &   1.34 &  28.23 \\
		43 &   ABCF &   1.34 &   1.34 &   1.34 &   8.07 &   8.07 &   1.34 \\
		44 &   ABDE &   1.34 &   1.34 &   8.07 &   1.34 &   1.34 &   8.07 \\
		45 &   ABDF &   1.34 &   1.34 &   8.07 &   1.34 &   8.07 &   1.34 \\
		46 &   ABEF &   1.34 &   1.34 &   8.07 &   8.07 &   1.34 &   1.34 \\
		47 &   ACDE &   1.34 &   8.07 &   1.34 &   1.34 &   1.34 &   8.07 \\
		48 &   ACDF &   1.34 &   8.07 &   1.34 &   1.34 &   8.07 &   1.34 \\
		49 &   ACEF &   1.34 &   8.07 &   1.34 &   8.07 &   1.34 &   1.34 \\
		50 &   ADEF &   1.34 &   8.07 &   8.07 &   1.34 &   1.34 &   1.34 \\
		51 &   BCDE &   8.07 &   1.34 &   1.34 &   1.34 &   1.34 &   8.07 \\
		52 &   BCDF &   8.07 &   1.34 &   1.34 &   1.34 &   8.07 &   1.34 \\
		53 &   BCEF &   8.07 &   1.34 &   1.34 &   8.07 &   1.34 &   1.34 \\
		54 &   BDEF &   8.07 &   1.34 &   8.07 &   1.34 &   1.34 &   1.34 \\
		55 &   CDEF &   8.07 &  28.23 &   1.34 &   1.34 &   1.34 &   1.34 \\
		56 &  ABCDE &   2.20 &   2.20 &   2.20 &   2.20 &   2.20 &  13.20 \\
		57 &  ABCDF &   2.20 &   2.20 &   2.20 &   2.20 &  13.20 &   2.20 \\
		58 &  ABCEF &   2.20 &   2.20 &   2.20 &  13.20 &   2.20 &   2.20 \\
		59 &  ABDEF &   2.20 &   2.20 &  13.20 &   2.20 &   2.20 &   2.20 \\
		60 &  ACDEF &   2.20 &  13.20 &   2.20 &   2.20 &   2.20 &   2.20 \\
		61 &  BCDEF &  13.20 &   2.20 &   2.20 &   2.20 &   2.20 &   2.20 \\
		\bottomrule
	\end{tabular}
			
		 }
\caption{Normalized squared  distances with 0.2 noise}
\label{tab:appA4}
	\end{center}
\end{table}

\clearpage

\begin{table}
	\begin{center}
		\scalebox{0.7}{
			
\begin{tabular}{@{}lllll@{}}
	\toprule
	Topic               & Distance     & Topic-len & alpha & Votes counter \\ \midrule
	BCDEFGHIJKLMNOPQRST & 0.0          & 19        & 0     & 1             \\
	CDEFHKLMOQRS        & 1.2          & 12        & 0.1   & 1             \\
	CDEHKLMOQRS         & 2.398685266  & 11        & 0.2   & 1             \\
	CDEHKLMOQRS         & 3.498685266  & 11        & 0.3   & 2             \\
	CDEHLMQRS           & 4.52939137   & 9         & 0.4   & 1             \\
	PR                  & 5.318912745  & 2         & 0.5   & 1             \\
	PR                  & 5.518912745  & 2         & 0.6   & 2             \\
	PR                  & 5.718912745  & 2         & 0.7   & 3             \\
	PR                  & 5.918912745  & 2         & 0.8   & 4             \\
	PR                  & 6.118912745  & 2         & 0.9   & 5             \\
	PR                  & 6.318912745  & 2         & 1.0   & 6             \\
	PR                  & 6.518912745  & 2         & 1.1   & 7             \\
	PR                  & 6.718912745  & 2         & 1.2   & 8             \\
	PR                  & 6.918912745  & 2         & 1.3   & 9             \\
	PR                  & 7.118912745  & 2         & 1.4   & 10            \\
	PR                  & 7.318912745  & 2         & 1.5   & 11            \\
	PR                  & 7.518912745  & 2         & 1.6   & 12            \\
	PR                  & 7.718912745  & 2         & 1.7   & 13            \\
	PR                  & 7.918912745  & 2         & 1.8   & 14            \\
	PR                  & 8.118912745  & 2         & 1.9   & 15            \\
	PR                  & 8.318912745  & 2         & 2.0   & 16            \\
	PR                  & 8.518912745  & 2         & 2.1   & 17            \\
	PR                  & 8.718912745  & 2         & 2.2   & 18            \\
	PR                  & 8.918912745  & 2         & 2.3   & 19            \\
	PR                  & 9.118912745  & 2         & 2.4   & 20            \\
	PR                  & 9.318912745  & 2         & 2.5   & 21            \\
	PR                  & 9.518912745  & 2         & 2.6   & 22            \\
	PR                  & 9.718912745  & 2         & 2.7   & 23            \\
	PR                  & 9.918912745  & 2         & 2.8   & 24            \\
	PR                  & 10.118912745 & 2         & 2.9   & 25            \\
	PR                  & 10.318912745 & 2         & 3.0   & 26            \\
	PR                  & 10.518912745 & 2         & 3.1   & 27            \\
	PR                  & 10.718912745 & 2         & 3.2   & 28            \\
	PR                  & 10.918912745 & 2         & 3.3   & 29            \\
	PR                  & 11.118912745 & 2         & 3.4   & 30            \\
	PR                  & 11.318912745 & 2         & 3.5   & 31            \\
	PR                  & 11.518912745 & 2         & 3.6   & 32            \\
	PR                  & 11.718912745 & 2         & 3.7   & 33            \\
	PR                  & 11.918912745 & 2         & 3.8   & 34            \\
	PR                  & 12.118912745 & 2         & 3.9   & 35            \\
	PR                  & 12.318912745 & 2         & 4.0   & 36            \\
	PR                  & 12.518912745 & 2         & 4.1   & 37            \\
	PR                  & 12.718912745 & 2         & 4.2   & 38            \\
	PR                  & 12.918912745 & 2         & 4.3   & 39            \\
	PR                  & 13.118912745 & 2         & 4.4   & 40            \\
	PR                  & 13.318912745 & 2         & 4.5   & 41            \\
	PR                  & 13.518912745 & 2         & 4.6   & 42            \\
	PR                  & 13.718912745 & 2         & 4.7   & 43            \\
	PR                  & 13.918912745 & 2         & 4.8   & 44            \\
	PR                  & 14.118912745 & 2         & 4.9   & 45            \\
	PR                  & 14.318912745 & 2         & 5.0   & 46            \\
	PR                  & 14.518912745 & 2         & 5.1   & 47            \\
	PR                  & 14.718912745 & 2         & 5.2   & 48            \\
	PR                  & 14.918912745 & 2         & 5.3   & 49            \\
	E                   & 15.072566372 & 1         & 5.4   & 1             \\ \bottomrule
\end{tabular}	
			
		}
		\caption{Doc1 supervised by U-map Deep, Method 2}
		\label{aic-doc1}
	\end{center}
\end{table}

\begin{table}
	\begin{center}
		\scalebox{0.7}{
			
\begin{tabular}{@{}lllll@{}}
	\toprule
	Topic               & Distance     & Topic-len & Noise & Votes counter \\ \midrule
	BCDEFGHIJKLMNOPQRST & 0.0          & 19        & 0     & 1             \\
	CLOR                & 5.111268331  & 4         & 0.5   & 1             \\
	PR                  & 5.994475803  & 2         & 1.0   & 1             \\
	PR                  & 6.78557188   & 2         & 1.5   & 2             \\
	PR                  & 7.551027265  & 2         & 2.0   & 3             \\
	PR                  & 8.294086823  & 2         & 2.5   & 4             \\
	PR                  & 9.017470057  & 2         & 3.0   & 5             \\
	PR                  & 9.723473297  & 2         & 3.5   & 6             \\
	PR                  & 10.414048928 & 2         & 4.0   & 7             \\
	PR                  & 11.09086746  & 2         & 4.5   & 8             \\
	PR                  & 11.755366629 & 2         & 5.0   & 9             \\
	PR                  & 12.408790572 & 2         & 5.5   & 10            \\
	PR                  & 13.052221354 & 2         & 6.0   & 11            \\
	PR                  & 13.686604528 & 2         & 6.5   & 12            \\
	PR                  & 14.312770002 & 2         & 7.0   & 13            \\
	PR                  & 14.931449197 & 2         & 7.5   & 14            \\
	PR                  & 15.543289234 & 2         & 8.0   & 15            \\
	PR                  & 16.148864739 & 2         & 8.5   & 16            \\
	PR                  & 16.748687717 & 2         & 9.0   & 17            \\
	PR                  & 17.34321586  & 2         & 9.5   & 18            \\
	PR                  & 17.932859558 & 2         & 10.0  & 19            \\
	PR                  & 18.51798786  & 2         & 10.5  & 20            \\
	PR                  & 19.098933547 & 2         & 11.0  & 21            \\
	PR                  & 19.675997486 & 2         & 11.5  & 22            \\
	PR                  & 20.249452363 & 2         & 12.0  & 23            \\
	PR                  & 20.819545905 & 2         & 12.5  & 24            \\
	PR                  & 21.386503677 & 2         & 13.0  & 25            \\
	PR                  & 21.950531502 & 2         & 13.5  & 26            \\
	PR                  & 22.511817576 & 2         & 14.0  & 27            \\
	PR                  & 23.070534315 & 2         & 14.5  & 28            \\
	PR                  & 23.626839976 & 2         & 15.0  & 29            \\
	PR                  & 24.180880079 & 2         & 15.5  & 30            \\
	PR                  & 24.732788667 & 2         & 16.0  & 31            \\
	PR                  & 25.282689409 & 2         & 16.5  & 32            \\
	PR                  & 25.830696594 & 2         & 17.0  & 33            \\
	PR                  & 26.376915994 & 2         & 17.5  & 34            \\
	PR                  & 26.92144565  & 2         & 18.0  & 35            \\
	PR                  & 27.464376562 & 2         & 18.5  & 36            \\
	PR                  & 28.00579331  & 2         & 19.0  & 37            \\
	PR                  & 28.545774613 & 2         & 19.5  & 38            \\
	PR                  & 29.084393825 & 2         & 20.0  & 39            \\
	PR                  & 29.621719387 & 2         & 20.5  & 40            \\
	PR                  & 30.157815229 & 2         & 21.0  & 41            \\
	PR                  & 30.692741138 & 2         & 21.5  & 42            \\
	PR                  & 31.226553085 & 2         & 22.0  & 43            \\
	PR                  & 31.759303529 & 2         & 22.5  & 44            \\
	PR                  & 32.291041685 & 2         & 23.0  & 45            \\
	PR                  & 32.821813772 & 2         & 23.5  & 46            \\
	PR                  & 33.351663238 & 2         & 24.0  & 47            \\
	PR                  & 33.880630964 & 2         & 24.5  & 48            \\
	PR                  & 34.408755449 & 2         & 25.0  & 49            \\
	PR                  & 34.936072987 & 2         & 25.5  & 50            \\
	PR                  & 35.462617815 & 2         & 26.0  & 51            \\
	PR                  & 35.98842226  & 2         & 26.5  & 52            \\
	PR                  & 36.513516874 & 2         & 27.0  & 53            \\
	PR                  & 37.037930548 & 2         & 27.5  & 54            \\
	PR                  & 37.561690626 & 2         & 28.0  & 55            \\
	PR                  & 38.08482301  & 2         & 28.5  & 56            \\
	PR                  & 38.607352247 & 2         & 29.0  & 57            \\
	PR                  & 39.129301625 & 2         & 29.5  & 58            \\
	PR                  & 39.650693245 & 2         & 30.0  & 59            \\
	PR                  & 40.171548101 & 2         & 30.5  & 60            \\
	E                   & 40.672566372 & 1         & 31.0  & 1             \\ \bottomrule
\end{tabular}
			
		}
		\caption{Doc1 supervised by U-map Deep, Method 3}
		\label{noise-doc1}
	\end{center}
\end{table}
\section{\\Appendix}

\end{document}